\documentclass{article}

\PassOptionsToPackage{numbers,compress}{natbib}
\usepackage[preprint]{neurips_2026}

\usepackage[utf8]{inputenc}
\usepackage[T1]{fontenc}
\usepackage{hyperref}
\hypersetup{pdftitle={Unreasonable Confidence from Simulator Misspecification: Directional Robustness on the SPEED+ Sim-to-Real Benchmark}, pdfauthor={Nikolay Dolgov}}
\usepackage{url}
\usepackage{booktabs}
\usepackage{amsfonts}
\usepackage{amsmath}
\usepackage{nicefrac}
\usepackage{microtype}
\usepackage{xcolor}
\usepackage{graphicx}
\usepackage{enumitem}

\definecolor{mplblue}{HTML}{1F77B4}
\definecolor{mplorange}{HTML}{FF7F0E}
\definecolor{mplgreen}{HTML}{2CA02C}

\newcommand{\Erot}{E_{R}}
\newcommand{\Etrans}{E_{T}}

\title{Unreasonable Confidence from Simulator Misspecification: Directional Robustness on the SPEED+ Sim-to-Real Benchmark}

\author{%
  Nikolay Dolgov\\
  Independent Researcher\\
  \texttt{niko.dolgow@gmail.com}
}

\begin{document}

\maketitle

\begin{abstract}
Vision-based navigation systems for orbit must be trained almost entirely on synthetic
imagery, because real labelled images of a given spacecraft in flight rarely exist.
Benchmarks like SPEED+ are built for that situation: the training set comes from a
simulator, and the test sets are real images of a satellite mockup under two kinds of
lighting. That simulator approximates how light reflects off mylar and aluminium, and like
every scientific simulator it is misspecified. Its error is radiometric, since it misses the
specular highlights, hard shadows and sensor saturation common in real images. We probe the resulting synthetic-to-real gap, the central risk for such systems,
with a deliberately simple monocular pose estimator. In a $2{\times}2$ study of backbone
adaptation and photometric augmentation, trained on synthetic images and tested on both real
hardware-in-the-loop domains, we find three things. (1)~The model does not know it is
failing: regions conformally calibrated to 90\% on synthetic data cover only 9--15\% of
rotation errors on sunlamp and 21--38\% on lightbox, so a wrong pose carries the confidence of
a correct one. (2)~Transfer is asymmetric: scaling data and features improved synthetic
rotation error from 1.65 to 1.03\,rad but moved the harshly lit \emph{sunlamp} domain by only
0.05\,rad, under twice its seed s.d., whereas adapting the backbone's last stage carried
63--83\% of its extra synthetic gain to both real domains. (3)~Once the backbone
adapts, augmentation helps only in the direction it models: our photometric stack improved the
diffusely lit \emph{lightbox} domain from 1.17 to 0.99\,rad on every seed but moved sunlamp no
more than seed-to-seed variation, while with the backbone frozen it helped sunlamp more. Every
post-review experiment was pre-registered, and the code, per-image predictions and
pre-registration record are at \url{https://github.com/lokindoo/spacecraft-pose-estimation}.
\end{abstract}

\section{Introduction}
\label{sec:intro}

Machine learning models for spacecraft navigation are trained almost entirely on the output
of a simulator, then deployed against a physical system that the simulator only approximates.
Real images of a target spacecraft in orbit, labelled with its pose, essentially do not
exist. Estimators therefore learn from rendered images and meet real ones for the first time
during the mission. The SPEED+ benchmark~\citep{park2022speedplus} was built to make that
discrepancy measurable. It pairs ${\sim}60{,}000$ labelled synthetic training images of the
Tango spacecraft with ${\sim}9{,}500$ real images of a physical mockup, lit two ways:
\emph{lightbox}, diffuse light like Earth albedo, and \emph{sunlamp}, a collimated solar
simulator that gives harsh directional light. On its predecessor SPEED, whose real test
images show the same mockup, the top ten entries of the 2019 challenge degraded
$2.7$--$40\times$ in pose score from synthetic to real images (Table~\ref{tab:spec2019}: winner
$40\times$, runner-up $5.3\times$, median $8.5\times$)~\citep{kisantal2020speed}.

SPEED+'s synthetic images do not come from a physics simulation. They are rendered by an
OpenGL rasteriser, the Optical Stimulator camera emulator, which shades sunlit foil and metal
with an empirical formula chosen for speed rather than fidelity~\citep{park2022speedplus}.
That formula stands in for the physics of light that the renderer never computes, much as a
climate model uses simple rules for the clouds it cannot resolve, and it is where the
simulator's error lives. The error is therefore radiometric rather than geometric: ranges and
pose distributions match across domains, and only appearance differs. Real mylar and machined
metal reflect specularly, and a collimated lamp turns those reflections into hard shadow edges
and saturated pixels. Saturation is not a shading error at all but happens in the camera's
detector, which means the simulator gets the camera wrong as well as the scene. In these
terms a photometric augmentation stack is a \emph{discrepancy model}, an unvalidated guess at
how simulator and reality differ. It does not correct the simulator but spreads the training
images across the differences it guesses at. This paper is about what happens when that guess
is wrong: the model loses accuracy, and it also loses its calibration without any outward
sign.

We use SPEED+ as a controlled testbed and ask \emph{what actually transfers from simulation
to reality}. Our estimator is a direct-regression baseline, an ImageNet-pretrained ResNet-50
with a small head. We train it on synthetic images alone, in a $2{\times}2$ design of
\{backbone frozen, last stage trained\} $\times$ \{no augmentation, photometric
augmentation\}, and evaluate it on the full lightbox and sunlamp test sets. Every arm runs on
three seeds (for the frozen un-augmented arm, two of them retrain only the head), and
checkpoints are selected on synthetic validation images only. Our contributions:

\begin{itemize}[leftmargin=1.4em,itemsep=1pt,topsep=2pt]
  \item \textbf{The model does not know it is failing.} Split-conformal regions calibrated to
  90\% on held-out synthetic data keep that coverage in-domain. On sunlamp they cover only
  9--15\% of rotation errors, and test-time BatchNorm adaptation lifts that only to 11--20\%.
  On the weakest arms the attitude cone spans roughly half of all orientations, which keeps
  it valid but makes it meaningless.
  \item \textbf{Transfer is asymmetric.} With the backbone frozen, scaling data and features
  improved synthetic rotation error from 1.65 to 1.03\,rad, but sunlamp only from 2.04 to
  1.99\,rad, less than twice its seed s.d., and the same holds for either factor alone.
  Training layer 4 took synthetic error further, from 1.03 to 0.32\,rad, and 63--83\% of that
  extra gain carried across: lightbox went from 1.76 to 1.17\,rad and sunlamp from 1.99 to
  1.53.
  \item \textbf{Augmentation is directional.} We built our photometric stack in good faith to
  anticipate harsh real illumination, yet it models neither real domain well, and the
  \emph{diffuse} lightbox markedly less badly (Fig.~\ref{fig:crops}). With layer 4 adapting,
  it improved lightbox by ${\sim}0.18$\,rad on every seed while sunlamp moved no further than
  seed-to-seed variation, and with the backbone frozen the order reverses. Removing blur,
  noise and the glare blob does not resolve sunlamp either.
  \item \textbf{The revision is pre-registered and auditable.} Every experiment added after
  review had its prediction and decision rule committed before its result was opened, and
  five of those rules softened or qualified a claim of the workshop submission
  (App.~\ref{app:prereg}). The per-image predictions are public, and so is a script that
  checks every number in this paper against the released results.
\end{itemize}

This extended version adds what the workshop's five-page limit left out.
Section~\ref{sec:related} places the study among prior work, including the 2019 challenge's
per-entry results, and Sec.~\ref{sec:baseline} reports what it took to get a baseline worth
testing. The post-review experiments move from the appendix into the body, and every arm now
has three seeds. Section~\ref{sec:failures} shows what the real-domain failures look like,
Sec.~\ref{sec:bn} brings together the results on normalisation statistics, and
Sec.~\ref{sec:conformal} adds coverage by range.
Appendices~\ref{app:rotshape}--\ref{app:prereg} give the per-checkpoint analyses, the
reproducibility details and the full pre-registration record.

\section{Related work}
\label{sec:related}

Prior work on spacecraft pose has measured how far methods fall across the synthetic-to-real
gap. We hold one simple method fixed and ask a different question: which of its gains cross
the gap, and does its uncertainty survive it?

\paragraph{Spacecraft pose benchmarks and the gap.} SPEED and its 2019
challenge~\citep{kisantal2020speed} set up the problem: synthetic training images, and real
test images of a Tango mockup. The top ten entries degraded $2.7$--$40\times$ from synthetic
to real (Table~\ref{tab:spec2019}). The runner-up, EPFL\_cvlab, had the best real score, and
the best entry without PnP, pedro\_fairspace, finished third. SPEED+~\citep{park2022speedplus}
made the gap itself the object of study, with two real domains, one lit diffusely and one by
a collimated sun simulator, and SPEC2021~\citep{park2023spec2021} reports the competition held
on them. SPNv2~\citep{park2024spnv2}, the reference method on SPEED+, combines multi-task
representations with heavy style augmentation, the texture randomisation that
\citet{park2019towards} first used with keypoints and PnP.

\begin{table}[ht]
  \caption{SPEC2019 top ten on SPEED~\citep{kisantal2020speed}: SPEED score on the synthetic and real test sets, and their ratio. Median ratio $8.5\times$.}
  \label{tab:spec2019}
  \centering
  \small
  \begin{tabular}{rl cc c c}
    \toprule
    \# & Team & $S_\mathrm{syn}$ & $S_\mathrm{real}$ & Ratio & PnP \\
    \midrule
    1 & UniAdelaide & 0.0094 & 0.3752 & $39.9\times$ & yes \\
    2 & EPFL\_cvlab & 0.0215 & 0.1140 & $5.3\times$ & yes \\
    3 & pedro\_fairspace & 0.0571 & 0.1555 & $2.7\times$ & no \\
    4 & Team Platypus & 0.0703 & 1.7201 & $24.5\times$ & no \\
    5 & motokimura1 & 0.0758 & 0.6011 & $7.9\times$ & no \\
    6 & Magpies & 0.1393 & 1.2659 & $9.1\times$ & no \\
    7 & GabrielA & 0.2423 & 2.6209 & $10.8\times$ & no \\
    8 & stainsby & 0.3711 & 5.0004 & $13.5\times$ & no \\
    9 & VSI\_Feeney & 0.4658 & 1.5993 & $3.4\times$ & no \\
    10 & jblumenkamp & 0.8859 & 4.2418 & $4.8\times$ & yes \\
    \bottomrule
  \end{tabular}
\end{table}

\paragraph{Direct regression and keypoints with PnP.} Direct methods map an image straight to
a pose, as PoseNet~\citep{kendall2015posenet} did for camera relocalisation and
SPN~\citep{sharma2020spn} and URSO~\citep{proenca2020urso} did for spacecraft. Keypoint
pipelines instead regress 2D landmarks and solve for the pose geometrically. The 2019
winner~\citep{chen2019landmark} regressed landmark heatmaps and refined the pose by nonlinear
optimisation, and keypoint methods took the top two places, but the ranking was on synthetic
data. On real data the 2019 results do not separate the two classes: PnP entries in the top
ten degraded $4.8$--$40\times$, the others $2.7$--$24\times$. For regression the rotation
representation matters too, since quaternions cover $SO(3)$ twice and every rotation
representation in four or fewer dimensions is discontinuous~\citep{zhou2019continuity}. We
keep the quaternion and remove the sign ambiguity instead (Sec.~\ref{sec:setup}).

\paragraph{Sim-to-real for vision.} Domain
randomisation~\citep{tobin2017domainrand,tremblay2018training} varies rendering so widely that
the real world looks like one more variation, and style augmentation~\citep{jackson2019style}
randomises texture and colour by style transfer. Our photometric stack is the cheap end of
the same idea, and its blur and noise are standard corruptions~\citep{hendrycks2019benchmarking}.
Test-time adaptation of BatchNorm
statistics~\citep{li2018adabn,schneider2020improving,nado2020evaluating} uses unlabelled
target images, as does adversarial domain adaptation~\citep{ganin2016domain}, and we test the
first (Sec.~\ref{sec:bn}) but not the second. Across many models and shifts, out-of-distribution
accuracy is often close to a linear function of in-distribution
accuracy~\citep{miller2021accuracy}, and robustness beyond that line is
rare~\citep{taori2020measuring}, a question Figure~\ref{fig:transfer} asks of our
configurations.

\paragraph{Uncertainty under shift.} Split conformal
prediction~\citep{vovk2005algorithmic,lei2018distribution,angelopoulos2023gentle} guarantees
coverage in finite samples, on average over test points, as long as calibration and test data
are exchangeable, which is exactly what a synthetic-to-real shift breaks. Two extensions relax
the assumption, but neither fits here. Weighted conformal prediction~\citep{tibshirani2019covariate}
restores validity under covariate shift given the likelihood ratio between domains, which
needs domains that overlap. Adaptive conformal inference~\citep{gibbs2021adaptive} tracks
coverage online from feedback on its own errors, and a navigator in orbit has no labels to
give that feedback. Coverage conditional on a feature such as range is attainable only in
weaker forms~\citep{vovk2012conditional,barber2021limits}, so we measure it
(Sec.~\ref{sec:conformal}) rather than guarantee it. Predictive uncertainty is known to
degrade under dataset shift~\citep{ovadia2019trust}, and conformal prediction has been carried
to object pose through keypoints~\citep{yang2023object}. We measure what a synthetic
calibration is worth on real images.

\section{Setup}
\label{sec:setup}

\paragraph{Data.} From SPEED+'s synthetic domain we sample 28{,}000 training images from the
official training labels and 2{,}000 validation images from the official validation set, and
downscale them to $320{\times}512$ grayscale to fit the GPU available. The conformal analysis
(Sec.~\ref{sec:conformal}) draws two more disjoint sets of 1{,}000 from the official
validation set, excluding those 2{,}000: one to calibrate on and one to test in-domain. Both
real domains serve in full as test sets, lightbox (6{,}740 images) and sunlamp (2{,}791), and
no real image influences training, tuning or model selection. Range distributions match across
domains, so any extra difficulty on the real images comes from appearance, not from harder
poses.

\paragraph{Model and loss.} A ResNet-50~\citep{he2016resnet} pretrained on
ImageNet~\citep{deng2009imagenet} feeds its global-average-pooled 2048-d features to a
two-layer MLP head. The head regresses translation $\mathbf{t}$ (targets z-scored per
component) and a unit quaternion $\mathbf{q}$: direct pose
regression~\citep{kendall2015posenet,sharma2020spn,proenca2020urso}. The loss treats the two
separately, $\mathcal{L} = \mathrm{MSE}(\hat{\mathbf{t}}, \mathbf{t}) + \lambda\,(1 -
\langle\hat{\mathbf{q}},\mathbf{q}\rangle^{2})$, with $\lambda{=}1$ and $\hat{\mathbf{q}}$
$\ell_2$-normalised. Because $\mathbf{q}$ and $-\mathbf{q}$ encode the same rotation, naive
quaternion MSE gives the network contradictory targets~\citep{zhou2019continuity}. Two choices
remove that ambiguity: the squared inner product ignores sign, and the labels are
canonicalised to one hemisphere. Section~\ref{sec:baseline} shows what it cost us to find
this out.

\paragraph{The $2{\times}2$ arms.} In the two \emph{frozen} arms the backbone runs in inference
mode and keeps its ImageNet BatchNorm statistics, so only the head is trained. In the two
\emph{layer-4} arms the final ResNet stage is trained as well, so its BatchNorm statistics
follow the (augmented) synthetic training images. All arms share one configuration: Adam,
learning rate $10^{-3}$, batch 16 and 20 epochs, with a checkpoint every second epoch. Each arm
keeps the checkpoint with the best \emph{synthetic} validation rotation error, and each runs
on seeds 42, 43 and 44. A frozen backbone without augmentation sees the same inputs in every
epoch, so its pooled features can be computed once and cached, and seeds 43 and 44 of that
arm retrain the head on the cache. As a check, the published seed-42 head applied to the
cache reproduces its errors to within $0.0005$\,rad (App.~\ref{app:revision}).

\paragraph{Augmentation.} The photometric factor is a good-faith attempt to anticipate harsh
real illumination. It jitters brightness (0.65--1.5) and contrast ($\pm$0.5), applies a
log-uniform gamma in $[0.5,2]$, and adds a soft, clamped Gaussian ``sun glare'' blob ($\sigma
\in [10,50]$\,px), Gaussian blur and additive Gaussian noise, each stochastic component with
$p{=}0.5$, followed by random erasing. Table~\ref{tab:augparams} gives every parameter, and
Sec.~\ref{sec:directional} examines what the stack actually prepares the model for.

\begin{table}[ht]
  \caption{The photometric stack, in order. Stochastic components apply with $p{=}0.5$. Arm A drops blur and noise, and arm B also drops the glare blob.}
  \label{tab:augparams}
  \centering
  \footnotesize
  \begin{tabular}{l l c c}
    \toprule
    Component & Parameters & Arm A & Arm B \\
    \midrule
    Brightness / contrast jitter & factor $[0.65, 1.5]$ / $\pm 0.5$ & \checkmark & \checkmark \\
    Gamma & log-uniform in $[0.5, 2]$ & \checkmark & \checkmark \\
    Sun-glare blob ($p{=}0.5$) & clamped Gaussian, $\sigma \in [10, 50]$\,px, peak $[0.8, 2]$ & \checkmark & -- \\
    Gaussian blur ($p{=}0.5$) & kernel 7, $\sigma \in [0.1, 2]$ & -- & -- \\
    Additive Gaussian noise ($p{=}0.5$) & $\sigma \in [0.01, 0.08]$ & -- & -- \\
    Random erasing ($p{=}0.5$) & area $[0.02, 0.06]$ & \checkmark & \checkmark \\
    \bottomrule
  \end{tabular}
\end{table}

\paragraph{Metrics.} We track normalised translation error $\Etrans =
\lVert\hat{\mathbf{t}}-\mathbf{t}\rVert / \lVert\mathbf{t}\rVert$ and geodesic rotation error
$\Erot = 2\arccos\lvert\langle\hat{\mathbf{q}}, \mathbf{q}\rangle\rvert$. Their sum is the
official SPEED+ score $S = \Etrans + \Erot$~\citep{kisantal2020speed}, which we report as a
plain sum, whereas SPEC2021~\citep{park2023spec2021} also zeroes errors below a calibration
threshold. Rotation errors also need a reference point, which uniform guessing provides: if
orientations are uniformly random, the rotation angle has density $(1-\cos\theta)/\pi$, so
random guessing scores
$\mathbb{E}[\theta] = \pi/2 + \nicefrac{2}{\pi} \approx 2.207$\,rad, which we call the
\emph{chance level}.\footnote{SPEED+
orientations are not uniform on $SO(3)$, so we also check the level empirically. A single
constant pose, fitted on the 28{,}000 training labels alone and never on a test domain, scores
$\Erot = 2.176$ on synthetic val, $2.203$ on lightbox and $2.199$ on sunlamp ($\Etrans =
0.415$, $0.367$, $0.372$). That is at most $0.031$\,rad from the uniform value, so we use
$2.207$ throughout. A fitted constant is achievable but not provably optimal, so these scores
are an upper bound on the error of the best constant predictor. Appendix~\ref{app:constant}
gives the method.} We define \emph{skill} as $2.207 - \Erot$, how far an error sits below chance,
because it separates errors that look close. As raw errors, 1.67 and 2.07\,rad are neighbours,
but as skill they are 0.54 and 0.13\,rad, one meaningfully better than chance and the other
barely.

\paragraph{Hardware.} The eight published runs were trained on a free-tier Colab GPU. The
post-review runs (the ablation arms, the control, and frozen augmented seeds 43 and 44) were
trained with the released code on an Apple M5 GPU, and the cached-feature heads on its CPU, as
App.~\ref{app:repro} lists run by run.

\paragraph{Code and data.} SPEED+ comes from its official hosts. The code, configs, per-seed
results, per-image predictions and conformal analysis behind every number here are at
\url{https://github.com/lokindoo/spacecraft-pose-estimation}.

\section{Getting an honest baseline}
\label{sec:baseline}

Before the gap can be measured, the in-domain error has to mean what it says. Two problems
stood between our first model and that, and neither was a sim-to-real effect
(Table~\ref{tab:baseline}).

\emph{The double cover.} Our first model regressed all seven pose targets, z-scored, with
plain MSE. A quaternion and its negative encode the same rotation, so near-identical images
can carry targets on opposite sides of the sphere, and MSE pulls the prediction towards their
average: the dataset mean~\citep{zhou2019continuity}. After 50 epochs on 4{,}000 images the
training loss was still 0.892, barely below the 1.0 that predicting the mean of z-scored
targets scores. Canonicalising the labels to one hemisphere, with nothing else changed,
brought the training loss to 0.285 and the validation rotation error to 1.666\,rad. This bug
would appear even if no real image existed, because it belongs to the labels rather than to
the shift.

\emph{Memorisation.} The next step scaled to 28{,}000 images and pooled 2048-d features. The
first such run (seed 42) scores 0.654\,rad on its own training images, 0.40\,rad better than on
validation. Those 0.40\,rad are a generalisation gap, and the rest of its validation error is
what a head on frozen ImageNet features can fit at all. More data attacks the first part, and
adapting the backbone the second.

One lesson carries beyond this study: take a bundled change apart before crediting any piece
of it, as Section~\ref{sec:asymmetry} does for data and features.

\begin{table}[ht]
  \caption{The baseline in sequence: $\Erot$ (rad), on the 2{,}000-image synthetic validation split used everywhere else.}
  \label{tab:baseline}
  \centering
  \small
  \begin{tabular}{l c cc}
    \toprule
    Step & Synthetic (val) & Lightbox & Sunlamp \\
    \midrule
    0. 4k images, logits head, MSE on raw quaternions & -- & -- & -- \\
    1. + labels canonicalised to one hemisphere & 1.666 & 2.002 & 2.073 \\
    2. + sign-invariant rotation loss & -- & -- & -- \\
    3. + pooled 2048-d features & -- & -- & -- \\
    4. + 28k images (published frozen arm, seed 42) & 1.053 & 1.752 & 2.048 \\
    \bottomrule
    \multicolumn{4}{@{}l}{\scriptsize Step 0 recorded only its loss, and steps 2 and 3 have no saved score on this split.}\\
  \end{tabular}
\end{table}

\section{Synthetic gains transfer asymmetrically}
\label{sec:asymmetry}

\begin{table}[ht]
  \caption{Main results: mean $\pm$ s.d.\ over seeds \{42, 43, 44\}. $\Etrans$ range-normalised, $\Erot$ in rad (chance level 2.207). Checkpoints are picked by best synthetic validation $\Erot$.}
  \label{tab:results}
  \centering
  \footnotesize
  \setlength{\tabcolsep}{2.4pt}
  \begin{tabular}{ll cc cc cc}
    \toprule
    & & \multicolumn{2}{c}{Synthetic (val)} & \multicolumn{2}{c}{Lightbox} & \multicolumn{2}{c}{Sunlamp} \\
    \cmidrule(lr){3-4} \cmidrule(lr){5-6} \cmidrule(lr){7-8}
    Backbone & Aug. & $\Etrans$ & $\Erot$ & $\Etrans$ & $\Erot$ & $\Etrans$ & $\Erot$ \\
    \midrule
    frozen$^{\dagger}$ & -- & $0.066{\pm}0.001$ & $1.044{\pm}0.009$ & $0.150{\pm}0.001$ & $1.763{\pm}0.024$ & $0.254{\pm}0.008$ & $2.004{\pm}0.049$ \\
    frozen$^{\ddagger}$ & \checkmark & $0.073{\pm}0.002$ & $1.185{\pm}0.019$ & $0.121{\pm}0.005$ & $1.708{\pm}0.031$ & $0.304{\pm}0.024$ & $1.876{\pm}0.010$ \\
    layer 4 & -- & $0.036{\pm}0.002$ & $0.317{\pm}0.010$ & $0.169{\pm}0.027$ & $1.166{\pm}0.018$ & $0.311{\pm}0.036$ & $1.534{\pm}0.075$ \\
    layer 4 & \checkmark & $0.042{\pm}0.002$ & $0.322{\pm}0.027$ & $0.150{\pm}0.017$ & $\mathbf{0.987{\pm}0.010}$ & $0.355{\pm}0.081$ & $1.468{\pm}0.119$ \\
    \bottomrule
    \multicolumn{8}{@{}p{0.98\linewidth}@{}}{\scriptsize $^{\dagger}$Seed 42 trained end to end, and seeds 43, 44 retrain the head on cached features (App.~\ref{app:revision}). Table~\ref{tab:frozen} retrains all three, and the text quotes it. $^{\ddagger}$Seeds 43, 44 on an Apple M5. Per-seed rows: App.~\ref{app:perseed}.}\\
  \end{tabular}
\end{table}

Table~\ref{tab:results} gives the results of all four arms, but we start further back, from our
first frozen-backbone configuration: 4{,}000 training images, with the head
reading the backbone's 1{,}000-d ImageNet logits rather than its pooled features. Over three
seeds, each head retrained on cached features, it reaches $\Erot = 1.651$\,rad on the
2{,}000-image synthetic validation split, 1.962 on lightbox and 2.040 on sunlamp
(Table~\ref{tab:frozen}). Scaling to 28{,}000 images and 2048-d pooled features improves
synthetic rotation error by 0.620\,rad (1.651\,$\to$\,1.030), but sunlamp by only 0.054\,rad
(2.040\,$\to$\,1.986). That is less than twice the scaled head's seed s.d.\ on sunlamp
(0.031\,rad), the yardstick our pre-registered rule uses (App.~\ref{app:prereg}), while
lightbox absorbs roughly a third of the gain (1.962\,$\to$\,1.761). In terms of skill, the
scaled head is 1.18\,rad better than chance on synthetic images but keeps only 19\% of that on
sunlamp (38\% on lightbox), so under harsh illumination the improvement all but vanishes.

The same cells separate the two factors, as pre-registered (App.~\ref{app:revision}). Data and
features split the synthetic gain roughly evenly (0.33 and 0.29\,rad), and neither factor alone
moves sunlamp by more than twice its seed s.d. Two checks rule out the easy explanations: a
tuned head gains another 0.10\,rad on synthetic but nothing on the real domains, and
re-estimating every BatchNorm statistic on synthetic images, with the weights fixed, makes the
frozen head worse everywhere (Sec.~\ref{sec:bn}). The layer-4 gain below is therefore not a
matter of normalisation statistics alone.

\begin{table}[ht]
  \caption{Frozen-backbone heads retrained on cached features: $\Erot$ (rad), mean $\pm$ s.d.\ over three seeds, every one a head retrain. The published configuration is Table~\ref{tab:results}'s frozen arm, whose seed 42 was trained end to end instead.}
  \label{tab:frozen}
  \centering
  \small
  \begin{tabular}{l ccc}
    \toprule
    Configuration & Synthetic (val) & Lightbox & Sunlamp \\
    \midrule
    4k images, 1000-d logits & $1.651{\pm}0.024$ & $1.962{\pm}0.012$ & $2.040{\pm}0.051$ \\
    4k images, 2048-d pooled & $1.395{\pm}0.008$ & $1.872{\pm}0.016$ & $2.062{\pm}0.033$ \\
    28k images, 1000-d logits & $1.353{\pm}0.026$ & $1.878{\pm}0.015$ & $2.016{\pm}0.027$ \\
    28k images, 2048-d pooled (published config.) & $1.030{\pm}0.017$ & $1.761{\pm}0.026$ & $1.986{\pm}0.031$ \\
    28k, pooled, tuned (lr $3{\times}10^{-4}$, 60 ep.) & $0.929{\pm}0.014$ & $1.751{\pm}0.035$ & $2.001{\pm}0.025$ \\
    28k, pooled, synthetic-BN backbone & $1.394{\pm}0.002$ & $2.064{\pm}0.040$ & $2.107{\pm}0.010$ \\
    \bottomrule
  \end{tabular}
\end{table}

Training layer 4 behaves differently: against the same frozen head, synthetic $\Erot$ drops
1.03\,$\to$\,0.32 (Tables~\ref{tab:frozen} and~\ref{tab:results}), and this time the real
domains follow, sunlamp 1.99\,$\to$\,1.53 and lightbox 1.76\,$\to$\,1.17. Between 63\%
(sunlamp) and 83\% (lightbox) of the extra synthetic gain carries across. What produced a gain
therefore matters as much as its size, since more data for a head on frozen ImageNet features
helped little while adapting the representation helped a lot. Even so, the layer-4 arm without
augmentation keeps only 36\% of its skill on sunlamp (55\% on lightbox), and its SPEED+ score
degrades 3.8--5.2$\times$ ($S{=}0.353$ synthetic, 1.335 lightbox, 1.845 sunlamp). That sits
at the low end of the 2019 challenge's range, but it is not evidence of a smaller gap
(Sec.~\ref{sec:discussion}).

\begin{figure}[ht]
  \centering
  \includegraphics[width=0.85\linewidth]{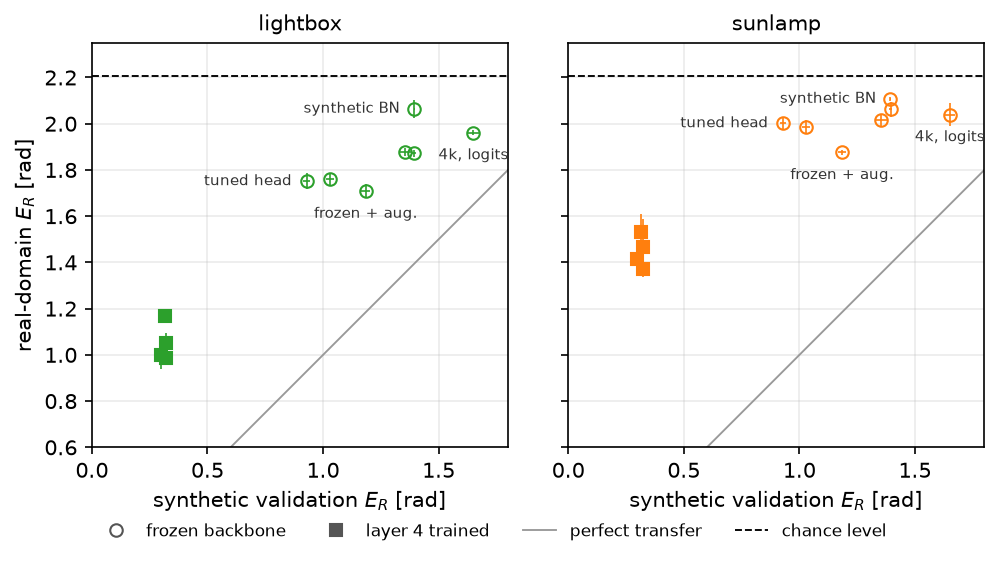}
  \caption{Synthetic against real rotation error for every configuration: frozen-backbone
  heads (open circles) and layer-4 arms (squares), mean $\pm$ s.d.\ over seeds. The grey line
  marks perfect transfer and the dashed line chance.}
  \label{fig:transfer}
\end{figure}

Figure~\ref{fig:transfer} puts the asymmetry in one picture, using the points of
Tables~\ref{tab:results} and~\ref{tab:frozen}. Across the six frozen configurations that keep ImageNet's BatchNorm statistics, synthetic error spans
0.93--1.65\,rad. Yet a least-squares line through them moves sunlamp error by only 0.11\,rad
per radian of synthetic error (lightbox 0.32). The layer-4 arms sit well off that line: at a
synthetic error of 0.30--0.32\,rad, every one of them beats every frozen configuration, by at
least 0.34\,rad on sunlamp and 0.54 on lightbox. In the terms of accuracy on the
line~\citep{miller2021accuracy}, the frozen configurations lie on a shallow line, and adapting
the backbone moves a configuration off it rather than along it.

\section{Augmentation is directional once the backbone adapts}
\label{sec:directional}

Which way augmentation helps depends on the backbone it trains. With layer 4 adapting, the
stack improves the diffuse domain it models and leaves sunlamp unresolved, but with the
backbone frozen the order reverses.

Seeds are paired across arms (42, 43 and 44 in both), so we read per-seed differences rather
than comparing spreads. On lightbox, augmentation lowers the error on every seed:
$-0.183$, $-0.185$ and $-0.170$\,rad, about 15\% each time, and the mean moves from
$1.166 \pm 0.018$ to $0.987 \pm 0.010$\,rad. On sunlamp the paired differences change sign:
$+0.116$, $-0.265$ and $-0.047$\,rad ($1.534 \pm 0.075 \to 1.468 \pm 0.119$), a spread wider
than the effect it would have to establish. Translation is a weaker indicator, but it points
the same way (Table~\ref{tab:results}). Lightbox improves within noise, while on sunlamp two
of the three augmented seeds ($0.405$, $0.399$) fail to beat the fitted constant's $0.372$, so
augmentation pushes sunlamp translation to the no-skill line.

With layer 4 adapting, the lightbox gain does not come at the cost of the synthetic
reference, whose error is essentially unchanged ($0.317 \to 0.322$\,rad). We report absolute error rather than the
synthetic-to-real ``gap'', because a gap can narrow merely because the reference gets worse.
With the backbone frozen, that is what happens: augmentation worsens synthetic error on every
seed (by $0.134$--$0.169$\,rad, paired against the head-only retrains in
Table~\ref{tab:frozenaug}). There the direction also reverses: sunlamp improves on every seed
($-0.069$ to $-0.132$\,rad), and by more than lightbox ($-0.012$ to $-0.117$), so augmentation
favours lightbox only once layer 4 adapts.

\begin{table}[ht]
  \caption{Frozen backbone with augmentation, $\Erot$ (rad) per seed and paired difference from the head-only retrain without augmentation (same seed).}
  \label{tab:frozenaug}
  \centering
  \small
  \setlength{\tabcolsep}{3pt}
  \begin{tabular}{c ccc ccc}
    \toprule
    & \multicolumn{3}{c}{$\Erot$} & \multicolumn{3}{c}{$\Delta\Erot$ vs.\ no augmentation} \\
    \cmidrule(lr){2-4} \cmidrule(lr){5-7}
    Seed & Synth. & Lightbox & Sunlamp & Synth. & Lightbox & Sunlamp \\
    \midrule
    42 & 1.180 & 1.718 & 1.864 & $+0.169$ & $-0.029$ & $-0.128$ \\
    43 & 1.205 & 1.734 & 1.883 & $+0.161$ & $-0.012$ & $-0.069$ \\
    44 & 1.169 & 1.673 & 1.881 & $+0.134$ & $-0.117$ & $-0.132$ \\
    \bottomrule
  \end{tabular}
\end{table}

\emph{Ablation} (Table~\ref{tab:ablation}). Dropping blur and noise lowers
sunlamp $\Erot$ on two of three seeds ($-0.184$, $+0.006$, $-0.117$\,rad) and worsens lightbox
on all three. Dropping the glare blob as well recovers lightbox, but not sunlamp. Neither
sunlamp effect is resolved: a control rerun on the same hardware shifts seed 42's sunlamp
error by $-0.142$\,rad, as much as the effects themselves.

\begin{table}[ht]
  \caption{Augmentation ablation, $\Erot$ (rad) per seed and paired difference from the published full stack (layer 4, same seed). Last column: sunlamp $\Etrans$.}
  \label{tab:ablation}
  \centering
  \small
  \setlength{\tabcolsep}{3pt}
  \begin{tabular}{l c ccc ccc c}
    \toprule
    & & \multicolumn{3}{c}{$\Erot$} & \multicolumn{3}{c}{$\Delta\Erot$ vs.\ full stack} & Sunlamp \\
    \cmidrule(lr){3-5} \cmidrule(lr){6-8}
    Arm & Seed & Synth. & Lightbox & Sunlamp & Synth. & Lightbox & Sunlamp & $\Etrans$ \\
    \midrule
    full stack & 42 & 0.352 & 0.986 & 1.592 & -- & -- & -- & 0.405 \\
     & 43 & 0.302 & 0.997 & 1.354 & -- & -- & -- & 0.399 \\
     & 44 & 0.311 & 0.977 & 1.459 & -- & -- & -- & 0.261 \\
    \midrule
    A: no blur/noise & 42 & 0.320 & 1.070 & 1.408 & $-0.032$ & $+0.084$ & $-0.184$ & 0.311 \\
     & 43 & 0.328 & 1.001 & 1.360 & $+0.026$ & $+0.004$ & $+0.006$ & 0.325 \\
     & 44 & 0.320 & 1.081 & 1.342 & $+0.009$ & $+0.104$ & $-0.117$ & 0.280 \\
    \midrule
    B: jitter, gamma, erasing & 42 & 0.297 & 0.983 & 1.390 & $-0.055$ & $-0.003$ & $-0.202$ & 0.286 \\
     & 43 & 0.310 & 0.947 & 1.437 & $+0.008$ & $-0.050$ & $+0.083$ & 0.309 \\
     & 44 & 0.291 & 1.068 & 1.414 & $-0.020$ & $+0.091$ & $-0.045$ & 0.330 \\
    \bottomrule
  \end{tabular}
\end{table}

\begin{figure}[ht]
  \centering
  \begin{minipage}[c]{0.47\linewidth}
    \includegraphics[width=\linewidth]{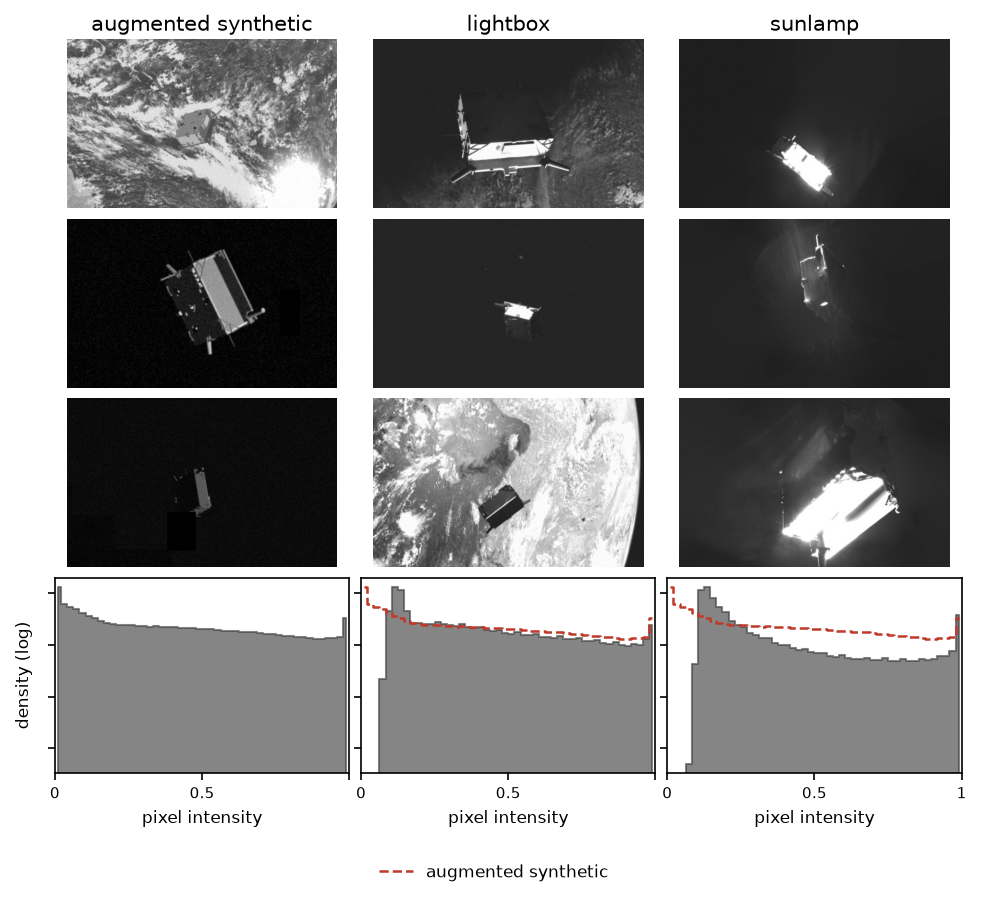}
  \end{minipage}\hfill
  \begin{minipage}[c]{0.49\linewidth}
    \caption{What the augmentation actually models. Columns: augmented synthetic training
  samples, and real test samples from each domain. Below each is the pixel-intensity density
  (log counts), pooled over every image in that split. The augmented density is overlaid on
  both real columns (dashed). It is far from both, and closer to lightbox: total variation
  $0.63$ against $0.80$. Gradient statistics give the same ordering
  (Table~\ref{tab:imagestats}).}
    \label{fig:crops}
  \end{minipage}
\end{figure}

\emph{Mechanism.} The stack models \emph{diffuse} illumination change: jitter and a soft,
clamped blob simulate gentle, spatially smooth variation in lighting, which is what lightbox
is. Sunlamp's collimated beam instead produces highlights off metal and foil, cast shadows, and
\emph{clipped} regions where the sensor saturates. Because photometric jitter only brightens
smoothly, the augmented images end up brighter overall than either real domain yet clip less
often than sunlamp (2.0\% of pixels against 2.8\%, Table~\ref{tab:imagestats}). We expected blur, noise and the glare blob
to work against sunlamp's sharp, bright structure. The ablation does not confirm this, and
per-pixel gradients do not even show sunlamp as sharper, since most of its frame is dark
(edge density 2.0\%, against 6.6\% for lightbox and 10.7\% for the augmented stack). What
gradients do show is direction: un-augmented synthetic images are closer to lightbox than to
sunlamp (gradient total variation 0.07 against 0.26), and the stack, mostly through blur and
noise, pushes them away from both (0.29 and 0.51). Layer 4's BatchNorm statistics follow these
augmented images, which fit neither real domain well and lightbox less badly
(Fig.~\ref{fig:crops}), so augmentation pushed training towards the domain it was already least
wrong about. Domain randomisation~\citep{tobin2017domainrand} helps where it brackets the
target domain, but ours brackets neither and misses sunlamp by more.

\begin{table}[ht]
  \caption{Pooled image statistics and total-variation distance to each real domain.}
  \label{tab:imagestats}
  \centering
  \small
  \begin{tabular}{l cc cc cc}
    \toprule
    & Edge & Clipped & \multicolumn{2}{c}{TV intensity} & \multicolumn{2}{c}{TV gradient} \\
    \cmidrule(lr){4-5} \cmidrule(lr){6-7}
    Pool & density (\%) & (\%) & lightbox & sunlamp & lightbox & sunlamp \\
    \midrule
    lightbox & 6.6 & 0.08 & 0.00 & 0.28 & 0.00 & 0.22 \\
    sunlamp & 2.0 & 2.78 & 0.28 & 0.00 & 0.22 & 0.00 \\
    synthetic, no aug. & 10.9 & 0.07 & 0.68 & 0.83 & 0.07 & 0.26 \\
    synthetic, full stack & 10.7 & 1.96 & 0.63 & 0.80 & 0.29 & 0.51 \\
    synthetic, arm A & 9.6 & 2.62 & 0.65 & 0.81 & 0.10 & 0.33 \\
    synthetic, arm B & 9.7 & 1.67 & 0.66 & 0.82 & 0.09 & 0.31 \\
    \bottomrule
  \end{tabular}
\end{table}

\section{What the failures look like}
\label{sec:failures}

The real-domain failures are broad rather than discrete confusions. A pose estimator can fail
in two quite different ways: it can mistake an object for a symmetric copy of itself, which
piles up errors near $\pi$, or it can spread its errors widely, and we pre-registered tests for
both (App.~\ref{app:prereg}).

\begin{figure}[t]
  \centering
  \includegraphics[width=0.8\linewidth]{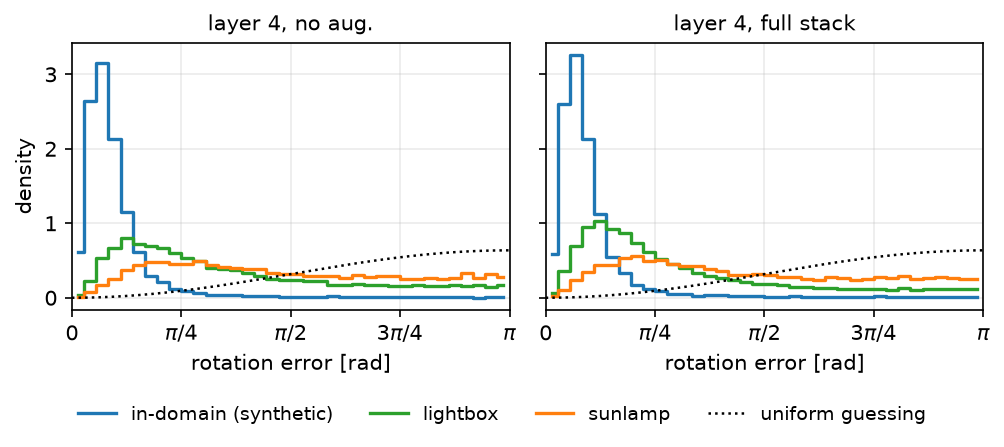}
  \caption{Rotation-error densities on each domain for the layer-4 arms, pooled over three
  seeds, against the density uniform guessing gives, $(1-\cos\theta)/\pi$ (dotted).}
  \label{fig:histograms}
\end{figure}

Neither real domain is bimodal (Fig.~\ref{fig:histograms}): on every layer-4 checkpoint the
share of errors within 0.3\,rad of $\pi$ is 0.03--0.05 on lightbox and 0.07--0.10 on sunlamp,
\emph{below} the 0.19 that uniform guessing puts there. Nor do the near-$\pi$ errors share an
axis. A symmetry flip would concentrate their axes in the body frame, but the share within
$20^{\circ}$ of the principal axis reaches the pre-registered 0.18 on only two of six
checkpoints per domain, and an in-plane confusion would concentrate them in the camera frame,
where the share reaches 0.18 on none (Table~\ref{tab:rotshape}). Lightbox errors peak near
$\pi/8$ (medians 0.69--0.92\,rad). Sunlamp errors spread from there towards $\pi$ (medians
1.11--1.54), with a long tail that stays below the uniform density beyond about $\pi/2$. Under
harsh light, then, the model is not mistaking Tango for a flipped copy of itself: its errors
are simply broad, even though its stated confidence says otherwise (Sec.~\ref{sec:conformal}).

\section{Normalisation statistics}
\label{sec:bn}

The layer-4 arms differ from the frozen ones in two ways at once: layer 4's weights adapt,
and so do its BatchNorm statistics, which then follow the augmented synthetic images. The
three results below pull these two changes apart.

First, statistics alone do not explain the layer-4 gain, since re-estimating every BatchNorm
layer of the frozen backbone on synthetic images, weights untouched, makes its head worse on
every domain: 1.394\,rad on synthetic, 2.064 on lightbox and 2.107 on sunlamp, against 1.030,
1.761 and 1.986 with ImageNet statistics (Table~\ref{tab:frozen}). Second, the direction
augmentation pushes in depends on whether layer 4, and with it its statistics, adapts. With
the backbone frozen, augmentation helps sunlamp more than lightbox on every seed, but once
layer 4 adapts it helps lightbox on every seed and leaves sunlamp unresolved
(Table~\ref{tab:frozenaug}). Third, swapping in real-domain statistics at test time helps
accuracy but not calibration. AdaBN~\citep{li2018adabn} re-estimates layer 4's statistics on
unlabelled images from one half of each real domain, and we test on the other half. It lowers
rotation error on every layer-4 checkpoint, by 0.09\,rad on lightbox and 0.16 on sunlamp on
average, yet rotation coverage reaches only 24--40\% and 11--20\% (Table~\ref{tab:adabn}). The
cone was calibrated on synthetic errors, and adaptation leaves it where it was.

\begin{table}[ht]
  \caption{AdaBN on the held-out half of each real domain: unadapted $\to$ adapted. Coverage is of the 90\% synthetic-calibrated rotation cone.}
  \label{tab:adabn}
  \centering
  \small
  \begin{tabular}{l c l ccc}
    \toprule
    Aug. & Seed & Domain & $\Erot$ & $\Etrans$ & Rotation coverage (\%) \\
    \midrule
    -- & 42 & lightbox & 1.171 $\to$ 1.074 & 0.202 $\to$ 0.149 & 21.1 $\to$ 23.9 \\
     &  & sunlamp & 1.519 $\to$ 1.393 & 0.298 $\to$ 0.196 & 8.5 $\to$ 11.2 \\
     & 43 & lightbox & 1.199 $\to$ 1.110 & 0.153 $\to$ 0.130 & 26.0 $\to$ 28.0 \\
     &  & sunlamp & 1.644 $\to$ 1.422 & 0.351 $\to$ 0.199 & 8.8 $\to$ 14.5 \\
     & 44 & lightbox & 1.160 $\to$ 1.057 & 0.156 $\to$ 0.132 & 25.6 $\to$ 29.3 \\
     &  & sunlamp & 1.540 $\to$ 1.393 & 0.283 $\to$ 0.181 & 11.7 $\to$ 11.6 \\
    \midrule
    \checkmark & 42 & lightbox & 1.012 $\to$ 0.933 & 0.153 $\to$ 0.136 & 36.4 $\to$ 40.4 \\
     &  & sunlamp & 1.594 $\to$ 1.400 & 0.406 $\to$ 0.210 & 12.3 $\to$ 18.6 \\
     & 43 & lightbox & 1.013 $\to$ 0.908 & 0.165 $\to$ 0.170 & 32.2 $\to$ 35.5 \\
     &  & sunlamp & 1.348 $\to$ 1.239 & 0.397 $\to$ 0.210 & 14.8 $\to$ 19.1 \\
     & 44 & lightbox & 0.978 $\to$ 0.914 & 0.133 $\to$ 0.129 & 37.5 $\to$ 40.0 \\
     &  & sunlamp & 1.483 $\to$ 1.342 & 0.262 $\to$ 0.192 & 14.6 $\to$ 19.6 \\
    \bottomrule
  \end{tabular}
\end{table}

In the discrepancy-model reading of Sec.~\ref{sec:intro}, the augmentation stack is a guess at
the simulator's error, and these results say where that guess enters the model. A frozen
backbone sees the guess only through what its head learns. An adapted one also absorbs it into
layer 4, normalisation included, and there the guess's bias towards diffuse light becomes the
model's own. Test-time adaptation replaces the guessed statistics with measured ones and
lowers the error, but it cannot repair a confidence that was calibrated under the guess.

AdaBN also lowers sunlamp translation error on every layer-4 checkpoint, by 0.07--0.20, to
0.18--0.21, against the fitted constant's 0.372. Without adaptation, predicted range on sunlamp
barely follows the true range (slope 0.11--0.33, against 0.98--1.01 on synthetic), which places
far targets much too close and is also why translation coverage collapses with range
(Sec.~\ref{sec:conformal}). No fixed position scores below 0.33 there, even one fitted to
sunlamp's own labels, so after adaptation the models read range from the image again.

\section{The model does not know it is failing}
\label{sec:conformal}

Does the estimator's uncertainty survive the shift? For each arm we calibrate split-conformal
regions~\citep{vovk2005algorithmic,lei2018distribution,angelopoulos2023gentle} on 1{,}000 fresh
synthetic images, used neither for training nor for checkpoint selection. For translation the
score is the error in metres, which gives a keep-out sphere, and for rotation it is the
geodesic angle, which gives a cone, both at a nominal 90\% with the finite-sample-corrected
quantile.

In-domain, the regions work: test coverage is close to nominal for every arm and metric
(87.7--91.4\%), and its Clopper--Pearson~\citep{clopper1934use} interval includes 90\% in every
case but one, the frozen un-augmented arm's translation (87.7\%, 85.5--89.7\%), which is
slightly anti-conservative.

\begin{figure}[ht]
  \centering
  \begin{minipage}[c]{0.37\linewidth}
    \includegraphics[width=\linewidth]{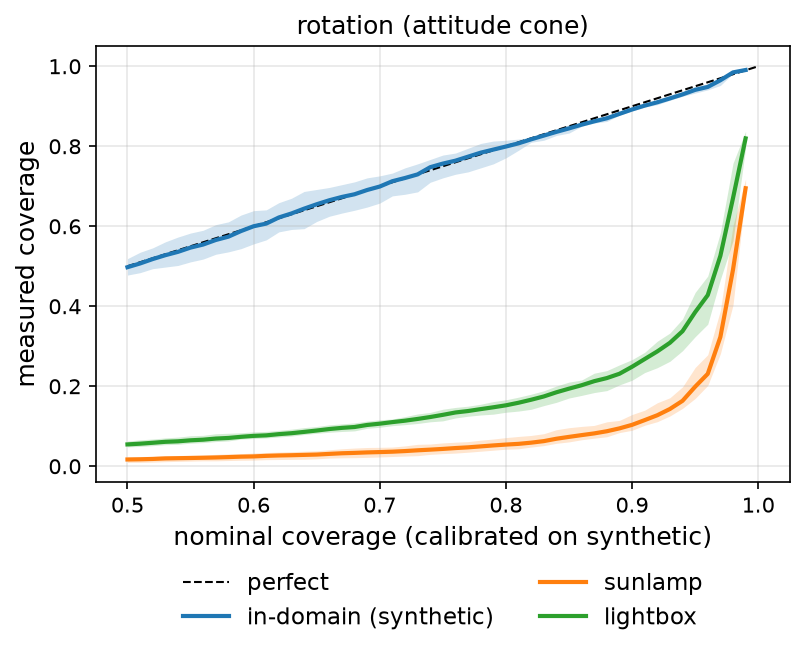}
  \end{minipage}\hfill
  \begin{minipage}[c]{0.59\linewidth}
    \caption{Conformal reliability of the attitude cone for the layer-4 arm without
  augmentation: mean over seeds 42, 43 and 44, with their range shaded. In-domain coverage
  tracks the diagonal, while both real domains collapse far below it. The guarantee itself is
  intact, and what broke is exchangeability.}
    \label{fig:reliability}
  \end{minipage}
\end{figure}

On the real domains coverage collapses (Table~\ref{tab:coverage-summary}): for the layer-4 arm without augmentation, over its
three seeds, rotation coverage is 9--13\% on sunlamp and 21--27\% on lightbox (translation
14--18\% and 33--35\%), and App.~\ref{app:conformal} gives every arm, domain and seed, with
intervals and inflation factors. To restore 90\% rotation coverage on sunlamp, that arm's
attitude cone would have to grow ${\sim}5$--$6\times$, from ${\sim}0.5$ to ${\sim}2.8$\,rad,
into a cone holding roughly 78\% of all orientations.\footnote{The fraction of $SO(3)$ lying
within geodesic angle $\alpha$ of a fixed rotation is $F(\alpha) = (\alpha - \sin\alpha)/\pi$,
so $F(2.79) \approx 0.78$, and $F(2.16)$ to $F(2.42)$ span $0.42$ to $0.56$.} Coverage follows
the directional finding of Sec.~\ref{sec:directional}: augmentation raises lightbox rotation
coverage to 33--38\%, but sunlamp only to 13--15\%. The frozen arms fail the opposite way,
because their 90\% attitude cone already spans 2.16--2.42\,rad, or 42--56\% of \emph{all
possible orientations}, which keeps it valid in-domain but makes it too wide to navigate by. A
navigation filter fed these
poses in deployment would receive wrong inputs that look exactly as confident as correct ones
(Fig.~\ref{fig:reliability}). Test-time adaptation does not rescue this
(Sec.~\ref{sec:bn}): AdaBN lowers the error, but lifts rotation coverage only to 24--40\% on
lightbox and 11--20\% on sunlamp.

\begin{table}[ht]
  \caption{Rotation coverage (\%) of the 90\% synthetic-calibrated cone, range over three seeds. AdaBN columns: after re-estimating layer 4's BatchNorm statistics on the other half of each real domain.}
  \label{tab:coverage-summary}
  \centering
  \footnotesize
  \setlength{\tabcolsep}{3pt}
  \begin{tabular}{l c ccc cc}
    \toprule
    & & & & & \multicolumn{2}{c}{With AdaBN} \\
    \cmidrule(lr){6-7}
    Arm & $\hat{q}_R$ (rad) & In-domain & Lightbox & Sunlamp & Lightbox & Sunlamp \\
    \midrule
    frozen, no aug. & 2.16--2.30 & 88.7--90.0 & 63.6--71.0 & 53.4--59.1 & -- & -- \\
    frozen, full stack & 2.37--2.42 & 88.7--91.1 & 75.1--76.3 & 69.8--71.0 & -- & -- \\
    layer 4, no aug. & 0.46--0.54 & 88.8--89.7 & 21.5--26.7 & 9.0--12.9 & 23.9--29.3 & 11.2--14.5 \\
    layer 4, full stack & 0.48--0.58 & 91.1--91.4 & 32.6--38.0 & 12.6--15.4 & 35.5--40.4 & 18.6--19.6 \\
    layer 4, arm A & 0.54--0.56 & 89.2--90.5 & 31.8--33.9 & 16.0--18.7 & -- & -- \\
    layer 4, arm B & 0.48--0.54 & 89.7--91.4 & 28.7--36.1 & 12.6--14.5 & -- & -- \\
    \bottomrule
  \end{tabular}
\end{table}

\begin{figure}[t]
  \centering
  \includegraphics[width=0.85\linewidth]{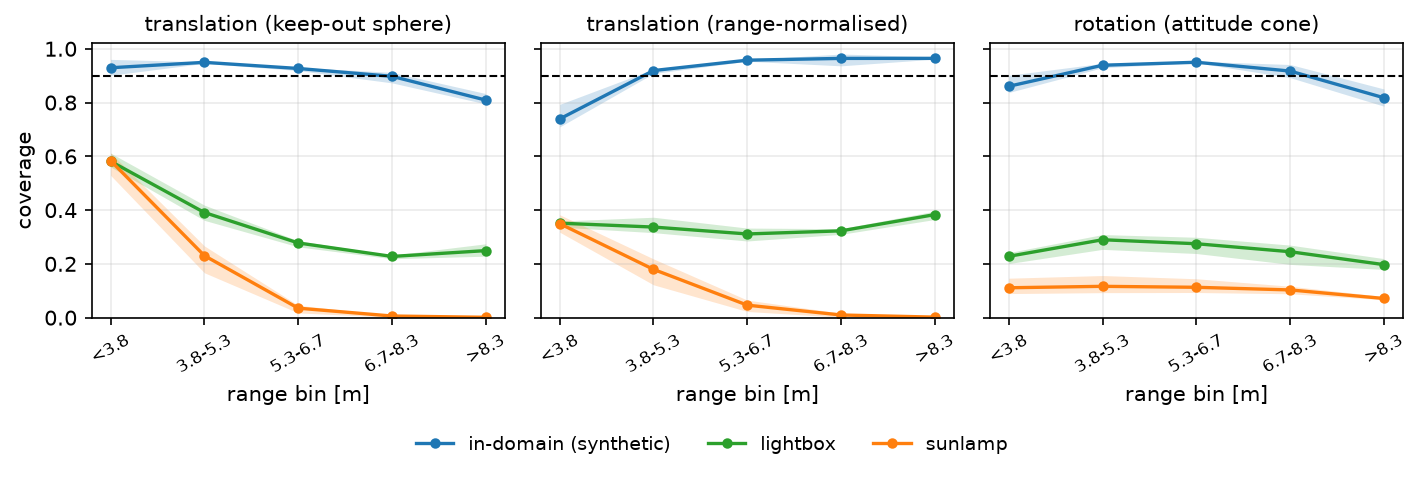}
  \caption{Coverage of the marginal 90\% region within each range bin (quintiles of the
  calibration split's range) for the layer-4 arm without augmentation: mean over three seeds,
  range shaded. Dashed: nominal.}
  \label{fig:range}
\end{figure}

\paragraph{Coverage by range.} Marginal coverage averages over poses, and in-domain it hides a
slope. For the layer-4 arm without augmentation, over three seeds, the metre-radius sphere
covers 92.9\% of the nearest fifth of in-domain targets but only 80.9\% of the farthest. The
range-normalised score reverses the slope: 73.9\% near and 96.4\% far (Fig.~\ref{fig:range}).
The rotation cone under-covers at both ends (86.0\% and 81.8\%) and over-covers in between. A
fixed metre radius is too tight far away and range-normalising overcorrects close in, so
neither score is conditionally calibrated, and a marginal guarantee cannot make it
so~\citep{barber2021limits}. The real domains are worse still, and on sunlamp translation
coverage collapses with range, from 58.4\% of the nearest fifth to 0.1\% of the farthest for the metre sphere, and from 35.0\% to 0.2\% when range-normalised, while
rotation coverage stays at 7.1--11.6\% throughout, and App.~\ref{app:range} gives every arm.

\section{Discussion and limitations}
\label{sec:discussion}

\paragraph{Ratios flatter weak models.} A degradation ratio divides by the in-domain error. On
sunlamp, even the skill-free constant pose (App.~\ref{app:constant}) scores only $7.3\times$
our synthetic $S$, so our modest ratio does not mean a small gap. Nor do the 2019 results
separate method classes: PnP entries in the top ten degraded $4.8$--$40\times$, the others
$2.7$--$24\times$ (Table~\ref{tab:spec2019}). SPNv2~\citep{park2024spnv2}, the reference
sim-to-real method on SPEED+, closes far more of the gap with multi-task representations and
heavy style augmentation. We do not compete with it but characterise how gains transfer under
a simple estimator.

\paragraph{Limitations.} In the frozen un-augmented arm, seed 42 was trained end to end and
seeds 43 and 44 retrain only the head on cached features. With the backbone frozen the two
compute the same thing, but they draw different random numbers and ran on different hardware:
when retrained head-only for comparison, seed 42 moves by 0.04\,rad on synthetic and 0.06 on
sunlamp, about as far as one seed from another. The difference is negligible, but we state it
because the frozen cell of Table~\ref{tab:results} combines the two. The ablation arms and two frozen augmented seeds
ran on different hardware from the published runs. Sunlamp is also the smaller test set
(2{,}791 images), and three seeds do not resolve sunlamp effects of 0.1\,rad. We study one
estimator on one benchmark at $320{\times}512$, and with real-domain errors near 1\,rad it is a
diagnostic baseline rather than a deployable navigator. Our directionality claim is about the
match between an augmentation and the shift it faces, not about photometric augmentation in
general. Our image statistics pool whole frames, which dark background dominates, so
object-masked statistics would test the mechanism more directly. Finally, our conformal
coverage is marginal, and Sec.~\ref{sec:conformal} shows how far it is from conditional.

\paragraph{Future work.} The mechanism suggests its own tests: augmentation that models
sunlamp's actual structure (specular highlights, hard shadows, saturation clipping), full
fine-tuning, and weighted conformal prediction~\citep{tibshirani2019covariate}. The last needs
density ratios between domains that barely overlap, which is exactly what sim-to-real lacks.

For fields that must rely on synthetic data, the question to ask of a robustness intervention
is not ``how much does it help?'' but ``which way does it push the training distribution?'',
and uncertainty has to be validated under the shift, or a wrong pose arrives wrapped in a
confident interval.

\bibliographystyle{plainnat}
\bibliography{references}

\begin{thebibliography}{32}
\providecommand{\natexlab}[1]{#1}
\providecommand{\url}[1]{\texttt{#1}}
\expandafter\ifx\csname urlstyle\endcsname\relax
  \providecommand{\doi}[1]{doi: #1}\else
  \providecommand{\doi}{doi: \begingroup \urlstyle{rm}\Url}\fi

\bibitem[Angelopoulos and Bates(2023)]{angelopoulos2023gentle}
Anastasios~N. Angelopoulos and Stephen Bates.
\newblock Conformal prediction: A gentle introduction.
\newblock \emph{Foundations and Trends in Machine Learning}, 16\penalty0 (4):\penalty0 494--591, 2023.

\bibitem[Chen et~al.(2019)Chen, Cao, Parra, and Chin]{chen2019landmark}
Bo~Chen, Jiewen Cao, Alvaro Parra, and Tat-Jun Chin.
\newblock Satellite pose estimation with deep landmark regression and nonlinear pose refinement.
\newblock In \emph{IEEE/CVF International Conference on Computer Vision Workshops (ICCVW)}, pages 2816--2824, 2019.

\bibitem[Clopper and Pearson(1934)]{clopper1934use}
Charles~J. Clopper and Egon~S. Pearson.
\newblock The use of confidence or fiducial limits illustrated in the case of the binomial.
\newblock \emph{Biometrika}, 26\penalty0 (4):\penalty0 404--413, 1934.

\bibitem[Deng et~al.(2009)Deng, Dong, Socher, Li, Li, and Fei-Fei]{deng2009imagenet}
Jia Deng, Wei Dong, Richard Socher, Li-Jia Li, Kai Li, and Li~Fei-Fei.
\newblock {ImageNet}: A large-scale hierarchical image database.
\newblock In \emph{IEEE Conference on Computer Vision and Pattern Recognition (CVPR)}, pages 248--255, 2009.

\bibitem[Foygel~Barber et~al.(2021)Foygel~Barber, Cand{\`e}s, Ramdas, and Tibshirani]{barber2021limits}
Rina Foygel~Barber, Emmanuel~J. Cand{\`e}s, Aaditya Ramdas, and Ryan~J. Tibshirani.
\newblock The limits of distribution-free conditional predictive inference.
\newblock \emph{Information and Inference: A Journal of the IMA}, 2021.

\bibitem[Ganin et~al.(2016)Ganin, Ustinova, Ajakan, Germain, Larochelle, Laviolette, Marchand, and Lempitsky]{ganin2016domain}
Yaroslav Ganin, Evgeniya Ustinova, Hana Ajakan, Pascal Germain, Hugo Larochelle, Fran{\c{c}}ois Laviolette, Mario Marchand, and Victor Lempitsky.
\newblock Domain-adversarial training of neural networks.
\newblock \emph{Journal of Machine Learning Research}, 17\penalty0 (59):\penalty0 1--35, 2016.

\bibitem[Gibbs and Cand{\`e}s(2021)]{gibbs2021adaptive}
Isaac Gibbs and Emmanuel Cand{\`e}s.
\newblock Adaptive conformal inference under distribution shift.
\newblock In \emph{Advances in Neural Information Processing Systems (NeurIPS)}, 2021.

\bibitem[He et~al.(2016)He, Zhang, Ren, and Sun]{he2016resnet}
Kaiming He, Xiangyu Zhang, Shaoqing Ren, and Jian Sun.
\newblock Deep residual learning for image recognition.
\newblock In \emph{IEEE Conference on Computer Vision and Pattern Recognition (CVPR)}, pages 770--778, 2016.

\bibitem[Hendrycks and Dietterich(2019)]{hendrycks2019benchmarking}
Dan Hendrycks and Thomas Dietterich.
\newblock Benchmarking neural network robustness to common corruptions and perturbations.
\newblock In \emph{International Conference on Learning Representations (ICLR)}, 2019.

\bibitem[Jackson et~al.(2019)Jackson, Atapour-Abarghouei, Bonner, Breckon, and Obara]{jackson2019style}
Philip~T. Jackson, Amir Atapour-Abarghouei, Stephen Bonner, Toby~P. Breckon, and Boguslaw Obara.
\newblock Style augmentation: Data augmentation via style randomization.
\newblock In \emph{IEEE/CVF Conference on Computer Vision and Pattern Recognition Workshops (CVPRW)}, 2019.

\bibitem[Kendall et~al.(2015)Kendall, Grimes, and Cipolla]{kendall2015posenet}
Alex Kendall, Matthew Grimes, and Roberto Cipolla.
\newblock {PoseNet}: A convolutional network for real-time {6-DOF} camera relocalization.
\newblock In \emph{IEEE International Conference on Computer Vision (ICCV)}, pages 2938--2946, 2015.

\bibitem[Kisantal et~al.(2020)Kisantal, Sharma, Park, Izzo, M{\"a}rtens, and D'Amico]{kisantal2020speed}
Mate Kisantal, Sumant Sharma, Tae~Ha Park, Dario Izzo, Marcus M{\"a}rtens, and Simone D'Amico.
\newblock Satellite pose estimation challenge: Dataset, competition design, and results.
\newblock \emph{IEEE Transactions on Aerospace and Electronic Systems}, 56\penalty0 (5):\penalty0 4083--4098, 2020.

\bibitem[Lei et~al.(2018)Lei, G'Sell, Rinaldo, Tibshirani, and Wasserman]{lei2018distribution}
Jing Lei, Max G'Sell, Alessandro Rinaldo, Ryan~J. Tibshirani, and Larry Wasserman.
\newblock Distribution-free predictive inference for regression.
\newblock \emph{Journal of the American Statistical Association}, 113\penalty0 (523):\penalty0 1094--1111, 2018.

\bibitem[Li et~al.(2018)Li, Wang, Shi, Hou, and Liu]{li2018adabn}
Yanghao Li, Naiyan Wang, Jianping Shi, Xiaodi Hou, and Jiaying Liu.
\newblock Adaptive batch normalization for practical domain adaptation.
\newblock \emph{Pattern Recognition}, 80:\penalty0 109--117, 2018.

\bibitem[Miller et~al.(2021)Miller, Taori, Raghunathan, Sagawa, Koh, Shankar, Liang, Carmon, and Schmidt]{miller2021accuracy}
John Miller, Rohan Taori, Aditi Raghunathan, Shiori Sagawa, Pang~Wei Koh, Vaishaal Shankar, Percy Liang, Yair Carmon, and Ludwig Schmidt.
\newblock Accuracy on the line: On the strong correlation between out-of-distribution and in-distribution generalization.
\newblock In \emph{International Conference on Machine Learning (ICML)}, 2021.

\bibitem[Nado et~al.(2020)Nado, Padhy, Sculley, D'Amour, Lakshminarayanan, and Snoek]{nado2020evaluating}
Zachary Nado, Shreyas Padhy, D.~Sculley, Alexander D'Amour, Balaji Lakshminarayanan, and Jasper Snoek.
\newblock Evaluating prediction-time batch normalization for robustness under covariate shift, 2020.
\newblock arXiv:2006.10963.

\bibitem[Ovadia et~al.(2019)Ovadia, Fertig, Ren, Nado, Sculley, Nowozin, Dillon, Lakshminarayanan, and Snoek]{ovadia2019trust}
Yaniv Ovadia, Emily Fertig, Jie Ren, Zachary Nado, D.~Sculley, Sebastian Nowozin, Joshua~V. Dillon, Balaji Lakshminarayanan, and Jasper Snoek.
\newblock Can you trust your model's uncertainty? {E}valuating predictive uncertainty under dataset shift.
\newblock In \emph{Advances in Neural Information Processing Systems (NeurIPS)}, 2019.

\bibitem[Park and D'Amico(2024)]{park2024spnv2}
Tae~Ha Park and Simone D'Amico.
\newblock Robust multi-task learning and online refinement for spacecraft pose estimation across domain gap.
\newblock \emph{Advances in Space Research}, 73\penalty0 (11):\penalty0 5726--5740, 2024.

\bibitem[Park et~al.(2019)Park, Sharma, and D'Amico]{park2019towards}
Tae~Ha Park, Sumant Sharma, and Simone D'Amico.
\newblock Towards robust learning-based pose estimation of noncooperative spacecraft.
\newblock In \emph{AAS/AIAA Astrodynamics Specialist Conference}, 2019.
\newblock arXiv:1909.00392.

\bibitem[Park et~al.(2022)Park, M{\"a}rtens, Lecuyer, Izzo, and D'Amico]{park2022speedplus}
Tae~Ha Park, Marcus M{\"a}rtens, Gurvan Lecuyer, Dario Izzo, and Simone D'Amico.
\newblock {SPEED+}: Next-generation dataset for spacecraft pose estimation across domain gap.
\newblock In \emph{IEEE Aerospace Conference}, 2022.
\newblock Dataset: Zenodo record 5588480.

\bibitem[Park et~al.(2023)Park, M{\"a}rtens, Jawaid, Wang, Zhao, Chin, Izzo, and D'Amico]{park2023spec2021}
Tae~Ha Park, Marcus M{\"a}rtens, Mohsi Jawaid, Zi~Wang, Bo~Zhao, Tat-Jun Chin, Dario Izzo, and Simone D'Amico.
\newblock Satellite pose estimation competition 2021: Results and analyses.
\newblock \emph{Acta Astronautica}, 204:\penalty0 640--665, 2023.

\bibitem[Proen{\c{c}}a and Gao(2020)]{proenca2020urso}
Pedro~F. Proen{\c{c}}a and Yang Gao.
\newblock Deep learning for spacecraft pose estimation from photorealistic rendering.
\newblock In \emph{IEEE International Conference on Robotics and Automation (ICRA)}, pages 6007--6013, 2020.

\bibitem[Schneider et~al.(2020)Schneider, Rusak, Eck, Bringmann, Brendel, and Bethge]{schneider2020improving}
Steffen Schneider, Evgenia Rusak, Luisa Eck, Oliver Bringmann, Wieland Brendel, and Matthias Bethge.
\newblock Improving robustness against common corruptions by covariate shift adaptation.
\newblock In \emph{Advances in Neural Information Processing Systems (NeurIPS)}, 2020.

\bibitem[Sharma and D'Amico(2020)]{sharma2020spn}
Sumant Sharma and Simone D'Amico.
\newblock Neural network-based pose estimation for noncooperative spacecraft rendezvous.
\newblock \emph{IEEE Transactions on Aerospace and Electronic Systems}, 56\penalty0 (6):\penalty0 4638--4658, 2020.

\bibitem[Taori et~al.(2020)Taori, Dave, Shankar, Carlini, Recht, and Schmidt]{taori2020measuring}
Rohan Taori, Achal Dave, Vaishaal Shankar, Nicholas Carlini, Benjamin Recht, and Ludwig Schmidt.
\newblock Measuring robustness to natural distribution shifts in image classification.
\newblock In \emph{Advances in Neural Information Processing Systems (NeurIPS)}, 2020.

\bibitem[Tibshirani et~al.(2019)Tibshirani, Barber, Cand{\`e}s, and Ramdas]{tibshirani2019covariate}
Ryan~J. Tibshirani, Rina~Foygel Barber, Emmanuel Cand{\`e}s, and Aaditya Ramdas.
\newblock Conformal prediction under covariate shift.
\newblock In \emph{Advances in Neural Information Processing Systems (NeurIPS)}, 2019.

\bibitem[Tobin et~al.(2017)Tobin, Fong, Ray, Schneider, Zaremba, and Abbeel]{tobin2017domainrand}
Josh Tobin, Rachel Fong, Alex Ray, Jonas Schneider, Wojciech Zaremba, and Pieter Abbeel.
\newblock Domain randomization for transferring deep neural networks from simulation to the real world.
\newblock In \emph{IEEE/RSJ International Conference on Intelligent Robots and Systems (IROS)}, pages 23--30, 2017.

\bibitem[Tremblay et~al.(2018)Tremblay, Prakash, Acuna, Brophy, Jampani, Anil, To, Cameracci, Boochoon, and Birchfield]{tremblay2018training}
Jonathan Tremblay, Aayush Prakash, David Acuna, Mark Brophy, Varun Jampani, Cem Anil, Thang To, Eric Cameracci, Shaad Boochoon, and Stan Birchfield.
\newblock Training deep networks with synthetic data: Bridging the reality gap by domain randomization.
\newblock In \emph{IEEE/CVF Conference on Computer Vision and Pattern Recognition Workshops (CVPRW)}, 2018.

\bibitem[Vovk(2012)]{vovk2012conditional}
Vladimir Vovk.
\newblock Conditional validity of inductive conformal predictors.
\newblock In \emph{Asian Conference on Machine Learning (ACML)}, 2012.

\bibitem[Vovk et~al.(2005)Vovk, Gammerman, and Shafer]{vovk2005algorithmic}
Vladimir Vovk, Alexander Gammerman, and Glenn Shafer.
\newblock \emph{Algorithmic Learning in a Random World}.
\newblock Springer, 2005.

\bibitem[Yang and Pavone(2023)]{yang2023object}
Heng Yang and Marco Pavone.
\newblock Object pose estimation with statistical guarantees: Conformal keypoint detection and geometric uncertainty propagation.
\newblock In \emph{IEEE/CVF Conference on Computer Vision and Pattern Recognition (CVPR)}, 2023.

\bibitem[Zhou et~al.(2019)Zhou, Barnes, Lu, Yang, and Li]{zhou2019continuity}
Yi~Zhou, Connelly Barnes, Jingwan Lu, Jimei Yang, and Hao Li.
\newblock On the continuity of rotation representations in neural networks.
\newblock In \emph{IEEE/CVF Conference on Computer Vision and Pattern Recognition (CVPR)}, pages 5745--5753, 2019.

\end{thebibliography}

\newpage
\appendix

\section{Per-seed results}
\label{app:perseed}

Table~\ref{tab:perseed} lists every run behind Table~\ref{tab:results}: all 36
arm\,$\times$\,seed\,$\times$\,domain evaluations, with the hardware each run was trained on.
Seed 42 of the frozen un-augmented arm was trained end to end, and seeds 43 and 44 retrain its
head on cached features. With the backbone frozen and the inputs un-augmented, the head sees
the same features either way, and the published seed-42 head reproduces its errors on the
cache to within $0.0005$ (App.~\ref{app:revision}). Table~\ref{tab:frozen} instead retrains the
head for all three seeds, so its row for the same configuration differs slightly from this
arm's mean, and the body quotes Table~\ref{tab:frozen} because its comparisons are made there.
Each run
keeps the checkpoint with the best \emph{synthetic} validation rotation error, chosen without
reference to either real domain.

\begin{table}[ht]
  \caption{Per-seed results behind Table~\ref{tab:results}. $\Etrans$ range-normalised, $\Erot$ in rad (chance level 2.207). \emph{Head only}: the head retrained on cached frozen features.}
  \label{tab:perseed}
  \centering
  \footnotesize
  \begin{tabular}{ll c l l cc}
    \toprule
    Backbone & Aug. & Seed & Trained on & Domain & $\Etrans$ & $\Erot$ \\
    \midrule
    frozen & -- & 42 & Colab & synthetic (val) & 0.066 & 1.053 \\
    frozen & -- & 42 & Colab & lightbox & 0.151 & 1.752 \\
    frozen & -- & 42 & Colab & sunlamp & 0.253 & 2.048 \\
    frozen & -- & 43 & head only & synthetic (val) & 0.064 & 1.045 \\
    frozen & -- & 43 & head only & lightbox & 0.150 & 1.746 \\
    frozen & -- & 43 & head only & sunlamp & 0.263 & 1.952 \\
    frozen & -- & 44 & head only & synthetic (val) & 0.066 & 1.036 \\
    frozen & -- & 44 & head only & lightbox & 0.148 & 1.791 \\
    frozen & -- & 44 & head only & sunlamp & 0.246 & 2.013 \\
    \midrule
    frozen & \checkmark & 42 & Colab & synthetic (val) & 0.071 & 1.180 \\
    frozen & \checkmark & 42 & Colab & lightbox & 0.118 & 1.718 \\
    frozen & \checkmark & 42 & Colab & sunlamp & 0.306 & 1.864 \\
    frozen & \checkmark & 43 & M5 & synthetic (val) & 0.073 & 1.205 \\
    frozen & \checkmark & 43 & M5 & lightbox & 0.126 & 1.734 \\
    frozen & \checkmark & 43 & M5 & sunlamp & 0.328 & 1.883 \\
    frozen & \checkmark & 44 & M5 & synthetic (val) & 0.074 & 1.169 \\
    frozen & \checkmark & 44 & M5 & lightbox & 0.118 & 1.673 \\
    frozen & \checkmark & 44 & M5 & sunlamp & 0.279 & 1.881 \\
    \midrule
    layer 4 & -- & 42 & Colab & synthetic (val) & 0.039 & 0.306 \\
    layer 4 & -- & 42 & Colab & lightbox & 0.200 & 1.169 \\
    layer 4 & -- & 42 & Colab & sunlamp & 0.299 & 1.476 \\
    layer 4 & -- & 43 & Colab & synthetic (val) & 0.035 & 0.326 \\
    layer 4 & -- & 43 & Colab & lightbox & 0.152 & 1.182 \\
    layer 4 & -- & 43 & Colab & sunlamp & 0.352 & 1.619 \\
    layer 4 & -- & 44 & Colab & synthetic (val) & 0.035 & 0.318 \\
    layer 4 & -- & 44 & Colab & lightbox & 0.154 & 1.147 \\
    layer 4 & -- & 44 & Colab & sunlamp & 0.283 & 1.506 \\
    \midrule
    layer 4 & \checkmark & 42 & Colab & synthetic (val) & 0.041 & 0.352 \\
    layer 4 & \checkmark & 42 & Colab & lightbox & 0.152 & 0.986 \\
    layer 4 & \checkmark & 42 & Colab & sunlamp & 0.405 & 1.592 \\
    layer 4 & \checkmark & 43 & Colab & synthetic (val) & 0.044 & 0.302 \\
    layer 4 & \checkmark & 43 & Colab & lightbox & 0.165 & 0.997 \\
    layer 4 & \checkmark & 43 & Colab & sunlamp & 0.399 & 1.354 \\
    layer 4 & \checkmark & 44 & Colab & synthetic (val) & 0.042 & 0.311 \\
    layer 4 & \checkmark & 44 & Colab & lightbox & 0.132 & 0.977 \\
    layer 4 & \checkmark & 44 & Colab & sunlamp & 0.261 & 1.459 \\
    \bottomrule
  \end{tabular}
\end{table}

\section{Conformal coverage, all arms and domains}
\label{app:conformal}

Split-conformal regions are calibrated per arm on 1{,}000 held-out synthetic images, used
neither for training nor for checkpoint selection, at nominal 90\% coverage with the
finite-sample-corrected quantile. Coverage carries Clopper--Pearson 95\%
intervals~\citep{clopper1934use}. $\hat{q}$ is the calibrated region size: a keep-out sphere
radius for translation, a cone half-angle for rotation. \emph{Inflation} is the factor by which
$\hat{q}$ would have to grow to restore 90\% coverage on a domain, so it measures directly how
wrong the calibration is. It sits near $1$ in-domain and reaches $1.2$--$13\times$ on the real
domains, worst on the sharper layer-4 arms. A low factor on the frozen arms is not good news:
their cones are already too wide in-domain to be informative. There are two kinds of
in-domain row: \emph{in-domain test} rows are a further 1{,}000 fresh synthetic images, and
\emph{Synthetic (val)} rows are the 2{,}000-image validation split used for checkpoint
selection, kept as a second in-domain control that is not fully independent.\footnote{The camera-ready labelled
these two ``synthetic (val)'' and ``synthetic (selection split)''.}

Table~\ref{tab:conformal-abs} uses absolute scores: Euclidean metres for translation, geodesic
radians for rotation. Table~\ref{tab:conformal-rel} repeats the translation analysis under the
range-normalised score. Table~\ref{tab:conformal-revision} gives rotation coverage for the
ablation arms and for every frozen-backbone head.

\begin{table}[ht]
  \caption{Conformal coverage at nominal 90\%, absolute scores, for every run in Table~\ref{tab:results}. Coverage in \% with Clopper--Pearson 95\% intervals, $\hat{q}_T$ in metres and $\hat{q}_R$ in radians.}
  \label{tab:conformal-abs}
  \centering
  \scriptsize
  \setlength{\tabcolsep}{3pt}
  \begin{tabular}{ll c l ccc ccc}
    \toprule
    & & & & \multicolumn{3}{c}{Translation (sphere)} & \multicolumn{3}{c}{Rotation (cone)} \\
    \cmidrule(lr){5-7} \cmidrule(lr){8-10}
    Backbone & Aug. & Seed & Domain & Cov. & $\hat{q}_T$ & Infl. & Cov. & $\hat{q}_R$ & Infl. \\
    \midrule
    frozen & -- & 42 & in-domain test & 87.7 (85.5--89.7) & 0.712 & 1.11 & 90.0 (88.0--91.8) & 2.299 & 1.01 \\
    frozen & -- & 42 & synthetic (val) & 88.0 (86.5--89.4) & 0.712 & 1.08 & 90.5 (89.1--91.7) & 2.299 & 0.99 \\
    frozen & -- & 42 & lightbox & 52.1 (50.9--53.3) & 0.712 & 2.48 & 71.0 (69.9--72.1) & 2.299 & 1.24 \\
    frozen & -- & 42 & sunlamp & 29.2 (27.5--30.9) & 0.712 & 4.76 & 58.2 (56.3--60.0) & 2.299 & 1.28 \\
    frozen & -- & 43 & in-domain test & 90.7 (88.7--92.4) & 0.751 & 0.96 & 88.9 (86.8--90.8) & 2.215 & 1.04 \\
    frozen & -- & 43 & synthetic (val) & 90.5 (89.1--91.7) & 0.751 & 0.98 & 89.1 (87.7--90.5) & 2.215 & 1.03 \\
    frozen & -- & 43 & lightbox & 55.4 (54.2--56.6) & 0.751 & 2.32 & 67.0 (65.9--68.1) & 2.215 & 1.30 \\
    frozen & -- & 43 & sunlamp & 32.0 (30.3--33.8) & 0.751 & 4.95 & 59.1 (57.3--61.0) & 2.215 & 1.31 \\
    frozen & -- & 44 & in-domain test & 88.3 (86.1--90.2) & 0.704 & 1.05 & 88.7 (86.6--90.6) & 2.159 & 1.03 \\
    frozen & -- & 44 & synthetic (val) & 88.6 (87.1--90.0) & 0.704 & 1.06 & 88.4 (87.0--89.8) & 2.159 & 1.04 \\
    frozen & -- & 44 & lightbox & 53.4 (52.2--54.6) & 0.704 & 2.58 & 63.6 (62.4--64.7) & 2.159 & 1.33 \\
    frozen & -- & 44 & sunlamp & 30.6 (28.9--32.3) & 0.704 & 4.72 & 53.4 (51.6--55.3) & 2.159 & 1.36 \\
    \midrule
    frozen & \checkmark & 42 & in-domain test & 89.3 (87.2--91.1) & 0.793 & 1.01 & 88.7 (86.6--90.6) & 2.403 & 1.02 \\
    frozen & \checkmark & 42 & synthetic (val) & 89.9 (88.5--91.2) & 0.793 & 1.01 & 89.5 (88.0--90.8) & 2.403 & 1.02 \\
    frozen & \checkmark & 42 & lightbox & 68.3 (67.2--69.4) & 0.793 & 1.84 & 75.1 (74.1--76.2) & 2.403 & 1.19 \\
    frozen & \checkmark & 42 & sunlamp & 26.3 (24.6--27.9) & 0.793 & 5.36 & 71.0 (69.3--72.7) & 2.403 & 1.20 \\
    frozen & \checkmark & 43 & in-domain test & 89.7 (87.6--91.5) & 0.828 & 1.01 & 90.9 (88.9--92.6) & 2.415 & 0.98 \\
    frozen & \checkmark & 43 & synthetic (val) & 89.6 (88.2--91.0) & 0.828 & 1.01 & 88.9 (87.4--90.2) & 2.415 & 1.03 \\
    frozen & \checkmark & 43 & lightbox & 66.9 (65.7--68.0) & 0.828 & 1.91 & 75.5 (74.4--76.5) & 2.415 & 1.18 \\
    frozen & \checkmark & 43 & sunlamp & 23.4 (21.8--25.0) & 0.828 & 5.16 & 70.0 (68.3--71.7) & 2.415 & 1.20 \\
    frozen & \checkmark & 44 & in-domain test & 89.8 (87.8--91.6) & 0.839 & 1.00 & 91.1 (89.2--92.8) & 2.372 & 0.97 \\
    frozen & \checkmark & 44 & synthetic (val) & 89.8 (88.4--91.1) & 0.839 & 1.01 & 89.5 (88.1--90.8) & 2.372 & 1.01 \\
    frozen & \checkmark & 44 & lightbox & 71.1 (70.0--72.2) & 0.839 & 1.72 & 76.3 (75.2--77.3) & 2.372 & 1.18 \\
    frozen & \checkmark & 44 & sunlamp & 30.1 (28.4--31.9) & 0.839 & 4.69 & 69.8 (68.1--71.5) & 2.372 & 1.21 \\
    \midrule
    layer 4 & -- & 42 & in-domain test & 89.3 (87.2--91.1) & 0.410 & 1.04 & 88.8 (86.7--90.7) & 0.460 & 1.06 \\
    layer 4 & -- & 42 & synthetic (val) & 89.6 (88.2--90.9) & 0.410 & 1.01 & 87.2 (85.7--88.7) & 0.460 & 1.10 \\
    layer 4 & -- & 42 & lightbox & 34.1 (32.9--35.2) & 0.410 & 7.92 & 21.5 (20.5--22.5) & 0.460 & 5.52 \\
    layer 4 & -- & 42 & sunlamp & 18.2 (16.8--19.6) & 0.410 & 10.92 & 9.1 (8.1--10.3) & 0.460 & 6.07 \\
    layer 4 & -- & 43 & in-domain test & 90.2 (88.2--92.0) & 0.378 & 0.99 & 89.1 (87.0--91.0) & 0.540 & 1.04 \\
    layer 4 & -- & 43 & synthetic (val) & 89.2 (87.8--90.5) & 0.378 & 1.02 & 89.2 (87.8--90.5) & 0.540 & 1.05 \\
    layer 4 & -- & 43 & lightbox & 35.4 (34.3--36.6) & 0.378 & 6.41 & 26.7 (25.6--27.7) & 0.540 & 4.76 \\
    layer 4 & -- & 43 & sunlamp & 13.8 (12.6--15.2) & 0.378 & 13.42 & 9.0 (8.0--10.1) & 0.540 & 5.23 \\
    layer 4 & -- & 44 & in-domain test & 89.9 (87.9--91.7) & 0.368 & 1.01 & 89.7 (87.6--91.5) & 0.526 & 1.02 \\
    layer 4 & -- & 44 & synthetic (val) & 88.4 (87.0--89.8) & 0.368 & 1.05 & 90.5 (89.1--91.7) & 0.526 & 0.99 \\
    layer 4 & -- & 44 & lightbox & 32.9 (31.8--34.1) & 0.368 & 6.52 & 26.5 (25.4--27.5) & 0.526 & 4.71 \\
    layer 4 & -- & 44 & sunlamp & 18.0 (16.6--19.5) & 0.368 & 11.73 & 12.9 (11.7--14.2) & 0.526 & 5.30 \\
    \midrule
    layer 4 & \checkmark & 42 & in-domain test & 89.8 (87.8--91.6) & 0.425 & 1.01 & 91.2 (89.3--92.9) & 0.576 & 0.97 \\
    layer 4 & \checkmark & 42 & synthetic (val) & 89.8 (88.3--91.0) & 0.425 & 1.01 & 90.0 (88.7--91.3) & 0.576 & 1.00 \\
    layer 4 & \checkmark & 42 & lightbox & 44.3 (43.1--45.5) & 0.425 & 6.45 & 37.2 (36.1--38.4) & 0.576 & 3.82 \\
    layer 4 & \checkmark & 42 & sunlamp & 10.0 (8.9--11.2) & 0.425 & 12.79 & 12.6 (11.4--13.9) & 0.576 & 4.88 \\
    layer 4 & \checkmark & 43 & in-domain test & 90.4 (88.4--92.2) & 0.494 & 0.98 & 91.1 (89.2--92.8) & 0.483 & 0.96 \\
    layer 4 & \checkmark & 43 & synthetic (val) & 91.3 (90.0--92.5) & 0.494 & 0.95 & 89.8 (88.3--91.0) & 0.483 & 1.01 \\
    layer 4 & \checkmark & 43 & lightbox & 47.8 (46.6--49.0) & 0.494 & 6.15 & 32.6 (31.5--33.7) & 0.483 & 4.85 \\
    layer 4 & \checkmark & 43 & sunlamp & 10.7 (9.6--11.9) & 0.494 & 10.80 & 15.3 (14.0--16.7) & 0.483 & 5.59 \\
    layer 4 & \checkmark & 44 & in-domain test & 91.0 (89.1--92.7) & 0.460 & 0.98 & 91.4 (89.5--93.1) & 0.544 & 0.92 \\
    layer 4 & \checkmark & 44 & synthetic (val) & 91.0 (89.6--92.2) & 0.460 & 0.97 & 90.8 (89.4--92.0) & 0.544 & 0.96 \\
    layer 4 & \checkmark & 44 & lightbox & 51.0 (49.8--52.2) & 0.460 & 5.10 & 38.0 (36.9--39.2) & 0.544 & 4.19 \\
    layer 4 & \checkmark & 44 & sunlamp & 24.4 (22.9--26.1) & 0.460 & 9.21 & 15.4 (14.1--16.8) & 0.544 & 5.03 \\
    \bottomrule
  \end{tabular}
\end{table}

\begin{table}[ht]
  \caption{Conformal coverage at nominal 90\%, translation scored with the range-normalised error $\Etrans$.}
  \label{tab:conformal-rel}
  \centering
  \scriptsize
  \setlength{\tabcolsep}{4pt}
  \begin{tabular}{ll c l ccc}
    \toprule
    Backbone & Aug. & Seed & Domain & Coverage & $\hat{q}$ (relative) & Infl. \\
    \midrule
    frozen & -- & 42 & in-domain test & 89.4 (87.3--91.2) & 0.1189 & 1.01 \\
    frozen & -- & 42 & synthetic (val) & 89.3 (87.9--90.7) & 0.1189 & 1.03 \\
    frozen & -- & 42 & lightbox & 49.6 (48.4--50.8) & 0.1189 & 2.42 \\
    frozen & -- & 42 & sunlamp & 21.7 (20.2--23.3) & 0.1189 & 3.85 \\
    frozen & -- & 43 & in-domain test & 90.3 (88.3--92.1) & 0.1217 & 0.99 \\
    frozen & -- & 43 & synthetic (val) & 91.2 (89.9--92.4) & 0.1217 & 0.94 \\
    frozen & -- & 43 & lightbox & 51.7 (50.5--52.9) & 0.1217 & 2.38 \\
    frozen & -- & 43 & sunlamp & 23.0 (21.5--24.6) & 0.1217 & 3.94 \\
    frozen & -- & 44 & in-domain test & 89.9 (87.9--91.7) & 0.1176 & 1.01 \\
    frozen & -- & 44 & synthetic (val) & 89.4 (88.0--90.7) & 0.1176 & 1.03 \\
    frozen & -- & 44 & lightbox & 50.6 (49.4--51.8) & 0.1176 & 2.50 \\
    frozen & -- & 44 & sunlamp & 22.1 (20.5--23.7) & 0.1176 & 3.74 \\
    \midrule
    frozen & \checkmark & 42 & in-domain test & 88.9 (86.8--90.8) & 0.1253 & 1.05 \\
    frozen & \checkmark & 42 & synthetic (val) & 89.6 (88.2--90.9) & 0.1253 & 1.01 \\
    frozen & \checkmark & 42 & lightbox & 66.2 (65.0--67.3) & 0.1253 & 1.78 \\
    frozen & \checkmark & 42 & sunlamp & 15.7 (14.3--17.1) & 0.1253 & 4.20 \\
    frozen & \checkmark & 43 & in-domain test & 86.2 (83.9--88.3) & 0.1259 & 1.10 \\
    frozen & \checkmark & 43 & synthetic (val) & 88.5 (87.0--89.9) & 0.1259 & 1.04 \\
    frozen & \checkmark & 43 & lightbox & 61.7 (60.5--62.9) & 0.1259 & 1.94 \\
    frozen & \checkmark & 43 & sunlamp & 13.4 (12.1--14.7) & 0.1259 & 4.46 \\
    frozen & \checkmark & 44 & in-domain test & 88.6 (86.5--90.5) & 0.1306 & 1.04 \\
    frozen & \checkmark & 44 & synthetic (val) & 89.5 (88.1--90.8) & 0.1306 & 1.02 \\
    frozen & \checkmark & 44 & lightbox & 67.8 (66.7--68.9) & 0.1306 & 1.69 \\
    frozen & \checkmark & 44 & sunlamp & 19.1 (17.7--20.6) & 0.1306 & 3.82 \\
    \midrule
    layer 4 & -- & 42 & in-domain test & 91.0 (89.1--92.7) & 0.0787 & 0.96 \\
    layer 4 & -- & 42 & synthetic (val) & 90.2 (88.8--91.5) & 0.0787 & 0.99 \\
    layer 4 & -- & 42 & lightbox & 34.1 (32.9--35.2) & 0.0787 & 6.07 \\
    layer 4 & -- & 42 & sunlamp & 13.5 (12.2--14.8) & 0.0787 & 6.93 \\
    layer 4 & -- & 43 & in-domain test & 92.0 (90.1--93.6) & 0.0739 & 0.93 \\
    layer 4 & -- & 43 & synthetic (val) & 91.3 (90.0--92.5) & 0.0739 & 0.95 \\
    layer 4 & -- & 43 & lightbox & 35.5 (34.4--36.7) & 0.0739 & 4.70 \\
    layer 4 & -- & 43 & sunlamp & 9.0 (8.0--10.2) & 0.0739 & 8.07 \\
    layer 4 & -- & 44 & in-domain test & 89.7 (87.6--91.5) & 0.0678 & 1.01 \\
    layer 4 & -- & 44 & synthetic (val) & 89.2 (87.8--90.6) & 0.0678 & 1.03 \\
    layer 4 & -- & 44 & lightbox & 32.2 (31.1--33.3) & 0.0678 & 5.11 \\
    layer 4 & -- & 44 & sunlamp & 12.0 (10.9--13.3) & 0.0678 & 7.68 \\
    \midrule
    layer 4 & \checkmark & 42 & in-domain test & 88.8 (86.7--90.7) & 0.0814 & 1.07 \\
    layer 4 & \checkmark & 42 & synthetic (val) & 90.5 (89.1--91.7) & 0.0814 & 0.99 \\
    layer 4 & \checkmark & 42 & lightbox & 43.7 (42.5--44.9) & 0.0814 & 4.58 \\
    layer 4 & \checkmark & 42 & sunlamp & 5.2 (4.4--6.1) & 0.0814 & 7.99 \\
    layer 4 & \checkmark & 43 & in-domain test & 90.3 (88.3--92.1) & 0.0911 & 1.00 \\
    layer 4 & \checkmark & 43 & synthetic (val) & 90.8 (89.5--92.1) & 0.0911 & 0.98 \\
    layer 4 & \checkmark & 43 & lightbox & 46.3 (45.1--47.5) & 0.0911 & 4.48 \\
    layer 4 & \checkmark & 43 & sunlamp & 5.7 (4.9--6.6) & 0.0911 & 6.96 \\
    layer 4 & \checkmark & 44 & in-domain test & 90.5 (88.5--92.2) & 0.0894 & 0.99 \\
    layer 4 & \checkmark & 44 & synthetic (val) & 91.0 (89.7--92.2) & 0.0894 & 0.97 \\
    layer 4 & \checkmark & 44 & lightbox & 51.4 (50.2--52.6) & 0.0894 & 3.60 \\
    layer 4 & \checkmark & 44 & sunlamp & 18.7 (17.2--20.2) & 0.0894 & 5.74 \\
    \bottomrule
  \end{tabular}
\end{table}

\begin{table}[ht]
  \caption{Rotation coverage (\%) at nominal 90\% for the ablation arms (per seed) and the frozen-backbone heads (range over seeds 42--44), each calibrated on its own predictions. $\hat{q}_R$ in rad.}
  \label{tab:conformal-revision}
  \centering
  \scriptsize
  \setlength{\tabcolsep}{4pt}
  \begin{tabular}{l c cccc c}
    \toprule
    Run & Seed & In-domain test & Synthetic (val) & Lightbox & Sunlamp & $\hat{q}_R$ \\
    \midrule
    A: no blur/noise & 42 & 90.5 & 91.1 & 33.2 & 16.0 & 0.558 \\
     & 43 & 89.2 & 88.8 & 33.9 & 16.8 & 0.538 \\
     & 44 & 90.4 & 90.5 & 31.8 & 18.7 & 0.540 \\
    \midrule
    B: jitter, gamma, erasing & 42 & 89.7 & 89.0 & 30.7 & 14.5 & 0.478 \\
     & 43 & 91.4 & 90.0 & 36.1 & 14.4 & 0.539 \\
     & 44 & 90.5 & 90.2 & 28.7 & 12.6 & 0.481 \\
    \midrule
    heads: 4k, logits & 42--44 & 89.1--92.4 & 89.5--92.2 & 85.3--87.1 & 83.0--85.3 & 2.81--2.86 \\
    heads: 4k, pooled & 42--44 & 88.5--91.9 & 88.5--92.5 & 77.0--83.2 & 71.0--77.7 & 2.57--2.71 \\
    heads: 28k, logits & 42--44 & 87.7--91.4 & 89.8--92.0 & 74.3--80.4 & 67.5--76.9 & 2.50--2.67 \\
    heads: 28k, pooled & 42--44 & 88.7--91.8 & 88.4--92.0 & 63.6--72.2 & 53.4--64.5 & 2.16--2.35 \\
    heads: tuned & 42--44 & 88.6--90.4 & 89.5--92.0 & 64.1--72.7 & 52.0--61.8 & 2.15--2.31 \\
    heads: synthetic BN & 42--44 & 89.4--90.1 & 89.3--90.5 & 73.6--77.1 & 72.2--74.2 & 2.64--2.67 \\
    \bottomrule
  \end{tabular}
\end{table}

\clearpage
\section{Constant-predictor baseline}
\label{app:constant}

The chance level of $\pi/2 + \nicefrac{2}{\pi} \approx 2.207$\,rad assumes orientations drawn
uniformly from $SO(3)$, which SPEED+ poses are not. This appendix gives the empirical
counterpart quoted in Section~\ref{sec:setup}: the score of a single fixed pose. It anchors
``near chance'' to a predictor one could actually build, rather than to an assumption about
the distribution.

The constant is fitted on the 28{,}000 training labels only. Fitting it on a test domain
would make it an oracle rather than a baseline, so neither real domain enters the estimate.
For rotation we take the principal eigenvector of $M = \sum_i \mathbf{q}_i
\mathbf{q}_i^{\top}$, which minimises chordal distance and is unaffected by the quaternion
double cover. We then refine it with five Karcher iterations on the sphere: sign-align each
$\mathbf{q}_i$ to the current mean, log-map into the tangent space, average, and exp-map back.
The $SO(3)$ geodesic distance is twice the $S^3$ one, so iterating on the sphere targets the
same minimiser as our reported metric. For translation we take the arithmetic mean of the
training translations. This gives $\mathbf{q} = (-0.198, 0.034, -0.019, -0.979)$ up to sign
and $\mathbf{t} = (-0.0001, 0.0026, 6.120)$\,m.

Two caveats belong with the numbers. First, the Karcher steps do not contract sharply here,
because the training orientations cover most of the sphere, which leaves the Fr\'echet
functional close to flat and the mean weakly identified. This does not affect the result: the
chordal and
Karcher estimates lie $0.022$\,rad apart and agree to $0.001$ on every score in
Table~\ref{tab:constant}. Second, both estimators minimise \emph{squared} error, whereas we
report mean unsquared error, whose minimisers are the corresponding geometric medians. The
scores below are therefore achievable rather than optimal, and they are an upper bound on the
error of the best constant predictor. We do not claim optimality, and no conclusion in the
paper needs it.

\begin{table}[ht]
  \caption{A constant pose fitted on the training split, evaluated on each test
  domain. $\Etrans$ range-normalised, $\Erot$ in radians. Skill is
  $2.207 - \Erot$. Reported for the Karcher-refined estimate.}
  \label{tab:constant}
  \centering
  \small
  \begin{tabular}{l c cc c}
    \toprule
    Domain & $n$ & $\Etrans$ & $\Erot$ & Skill \\
    \midrule
    synthetic (val) & 2{,}000 & 0.415 & 2.176 & 0.031 \\
    lightbox & 6{,}740 & 0.367 & 2.203 & 0.005 \\
    sunlamp & 2{,}791 & 0.372 & 2.199 & 0.009 \\
    \bottomrule
  \end{tabular}
\end{table}

The empirical no-skill rotation level therefore sits at most $0.031$\,rad from the uniform
one on all three domains, and we use $2.207$ in the body. Every trained run behind
Table~\ref{tab:results} keeps some skill over it, even the weakest, seed 42 of the frozen
un-augmented arm on sunlamp, which scores $2.048$\,rad and keeps $0.15$\,rad.

Translation is less comfortable, and it is where the second caveat matters most. On sunlamp
the constant scores $0.372$, which two of the three augmented layer-4 seeds ($0.405$, $0.399$)
do not beat, while a constant fitted directly to the unsquared metric does better on
translation, scoring
$0.330$ on sunlamp, and matches this one on rotation to within $0.001$. We quote the
Karcher-refined constant throughout, as the workshop version did, because no conclusion
changes with the tighter one: the same two seeds fail to beat it, and every other comparison
the paper draws with the constant still holds.

\section{Revision experiments}
\label{app:revision}

All runs in this appendix were added after review and pre-registered: predictions and
decision rules were written down and committed before any result was opened
(App.~\ref{app:prereg}), and selection uses synthetic validation only, as in the body.

\paragraph{Frozen-backbone block.} The frozen backbone runs in inference mode, so without
augmentation its pooled features (and the ImageNet logits computed from them) can be cached
once, and heads retrained on them in minutes. Applying the published seed-42 head to the
cache reproduces its published errors to within $0.0005$. Table~\ref{tab:frozen} gives three
seeds (42, 43, 44) for each configuration, all with the published head, loss and selection
rule. The 4{,}000-image cells use one fixed draw from the 28{,}000, not the original
4{,}000-image set. The tuned head is the best of learning rate $\{3{\times}10^{-4}, 10^{-3},
3{\times}10^{-3}\}\times\{20, 60\}$ epochs on seed 42, chosen on synthetic validation. The
synthetic-BN head re-estimates every BatchNorm layer on the un-augmented synthetic training
images, with the weights untouched, before caching.

\paragraph{Frozen backbone with augmentation.} Augmented inputs cannot use the feature cache,
so seeds 43 and 44 of the frozen augmented arm were trained end to end with the released code
on an Apple M5 GPU, while seed 42 is the published Colab run. Table~\ref{tab:frozenaug} pairs
each seed with the head-only retrain of the same seed in the published configuration
(Table~\ref{tab:frozen}), and their conformal coverage is in App.~\ref{app:conformal}.

\paragraph{Augmentation ablation.} Arm A is the full stack without blur and noise. Arm B keeps
only brightness and contrast jitter, gamma and random erasing (no glare blob, blur or noise).
Both train layer 4 on seeds 42, 43 and 44 with the published configuration, and we read them
as per-seed differences. They were trained with the released code on an Apple M5 GPU, whereas
the published arms were trained on Colab. A control rerun of the full stack (seed 42) on the
M5 differs from its published counterpart by $-0.053$, $+0.006$ and $-0.142$\,rad in $\Erot$
(synthetic, lightbox, sunlamp). That is within seed variation, but on sunlamp it is as large
as the ablation effects themselves.

\paragraph{Test-time BatchNorm adaptation (AdaBN).} For each layer-4 checkpoint we
re-estimate only layer 4's BatchNorm statistics, the ones fitted on synthetic data. We use
the unlabelled images of one half of each real domain (a fixed random split by filename),
leave every weight untouched, and evaluate on the other half (3{,}370 lightbox and 1{,}396
sunlamp images), with each arm's published synthetic-calibrated $\hat{q}$ for coverage. This is
test-time adaptation: it uses target images, though not their labels. The frozen arms are
excluded, as all their statistics are ImageNet's.

\paragraph{Image statistics.} Figure~\ref{fig:crops} compares pooled intensity densities,
which cannot tell a soft bright blob from a hard edge. Table~\ref{tab:imagestats} adds spatial
statistics, pooled over every image of each split after the same resize and grayscale
conversion (augmentation seed 7, as in Figure~\ref{fig:crops}):
\begin{itemize}[leftmargin=1.2em,itemsep=0pt,topsep=2pt]
  \item the Sobel gradient magnitude, scaled so a full black-to-white step reads 1
  \item edge density, the share of pixels whose gradient exceeds 0.1
  \item the clipped fraction, the share of pixels at or above $254/255$
  \item the total variation (TV) between pooled densities (48 intensity bins, 64 gradient
  bins)
\end{itemize}
The intensity TV of the full stack reproduces Figure~\ref{fig:crops} ($0.63$, $0.80$).

\section{Rotation-error shape}
\label{app:rotshape}

For each layer-4 checkpoint and real domain we measure two things, both pre-registered
(App.~\ref{app:prereg}). The first is the share of rotation errors within 0.3\,rad of $\pi$.
Under uniform guessing the angle density, $(1-\cos\theta)/\pi$, rises towards $\pi$, so chance
alone puts 0.19 of the mass there. The second is how concentrated the axes of those near-$\pi$
errors are, in the body frame ($R_\mathrm{true}^{\top} R_\mathrm{pred}$) and in the camera
frame ($R_\mathrm{pred} R_\mathrm{true}^{\top}$). We measure it as the share of axes within
$20^{\circ}$ of their principal axis, the leading eigenvector of the axes' second-moment
matrix. A flip about a near-symmetry axis would concentrate the body-frame axes, whereas an
in-plane confusion about the boresight would concentrate the camera-frame ones. Because the principal
axis is estimated from the same sample, the share's chance level is not the 0.06 of a fixed
axis. Simulation with uniform random axes gives 0.074--0.079 at these sample sizes (95th
percentile about 0.10).

\begin{table}[ht]
  \caption{Shape of the rotation errors on the real domains, layer-4 checkpoints. Near-$\pi$: share of errors within 0.3\,rad of $\pi$ (uniform guessing: 0.19). Cone: share of the near-$\pi$ errors' axes within $20^{\circ}$ of their principal axis, in the body and camera frames. A symmetry flip would concentrate the first and an in-plane confusion the second.}
  \label{tab:rotshape}
  \centering
  \scriptsize
  \setlength{\tabcolsep}{4pt}
  \begin{tabular}{l l c c c cc}
    \toprule
    Run & Domain & Median $\Erot$ & Near-$\pi$ & $n_{\pi}$ & Cone, body & Cone, camera \\
    \midrule
    layer 4, no aug., 42 & lightbox & 0.92 & 0.051 & 344 & 0.105 & 0.122 \\
     & sunlamp & 1.31 & 0.086 & 240 & 0.083 & 0.071 \\
    layer 4, no aug., 43 & lightbox & 0.92 & 0.050 & 339 & 0.198 & 0.112 \\
     & sunlamp & 1.54 & 0.095 & 265 & 0.132 & 0.068 \\
    layer 4, no aug., 44 & lightbox & 0.90 & 0.044 & 299 & 0.100 & 0.094 \\
     & sunlamp & 1.37 & 0.082 & 229 & 0.140 & 0.118 \\
    layer 4, full stack, 42 & lightbox & 0.74 & 0.027 & 183 & 0.191 & 0.131 \\
     & sunlamp & 1.50 & 0.090 & 250 & 0.128 & 0.092 \\
    layer 4, full stack, 43 & lightbox & 0.71 & 0.040 & 270 & 0.163 & 0.144 \\
     & sunlamp & 1.11 & 0.071 & 198 & 0.192 & 0.096 \\
    layer 4, full stack, 44 & lightbox & 0.69 & 0.039 & 263 & 0.171 & 0.122 \\
     & sunlamp & 1.31 & 0.071 & 197 & 0.315 & 0.071 \\
    \bottomrule
  \end{tabular}
\end{table}

\clearpage
\section{Coverage by range bin}
\label{app:range}

For each run we calibrate $\hat{q}$ on the calibration split exactly as in
Sec.~\ref{sec:conformal}, hold it fixed, and measure coverage within five range bins. The bin
edges are quintiles of the calibration split's range, so no test image shapes them. In-domain
bins hold 173--234 images each, and the per-bin counts sum to the marginal coverage on every
run. Tables~\ref{tab:rangebins-translation-metres}, \ref{tab:rangebins-translation-relative}
and~\ref{tab:rangebins-rotation} give each arm's coverage per bin, averaged over its seeds, for
the metre sphere, the range-normalised sphere and the attitude cone.

\begin{table}[ht]
  \caption{Coverage (\%) of the marginal 90\% region by range bin (m), translation, metres, as the mean over the seeds of each arm. Bins are quintiles of the calibration split's range.}
  \label{tab:rangebins-translation-metres}
  \centering
  \scriptsize
  \setlength{\tabcolsep}{4pt}
  \begin{tabular}{l l ccccc}
    \toprule
    Arm & Domain & $<3.8$ & 3.8--5.3 & 5.3--6.7 & 6.7--8.3 & $\geq8.3$ \\
    \midrule
    frozen, no aug. & in-domain & 97.5 & 95.7 & 90.2 & 82.9 & 75.2 \\
     & lightbox & 73.3 & 70.7 & 50.5 & 37.4 & 28.6 \\
     & sunlamp & 63.6 & 53.0 & 22.2 & 6.6 & 1.4 \\
    \midrule
    frozen, full stack & in-domain & 97.4 & 95.0 & 92.1 & 90.9 & 75.6 \\
     & lightbox & 91.3 & 84.1 & 64.4 & 58.2 & 45.6 \\
     & sunlamp & 81.7 & 39.4 & 11.6 & 3.0 & 0.1 \\
    \midrule
    layer 4, no aug. & in-domain & 92.9 & 95.0 & 92.7 & 89.8 & 80.9 \\
     & lightbox & 58.2 & 39.1 & 27.8 & 22.8 & 24.9 \\
     & sunlamp & 58.4 & 23.1 & 3.5 & 0.6 & 0.1 \\
    \midrule
    layer 4, full stack & in-domain & 92.0 & 94.1 & 93.6 & 89.3 & 84.8 \\
     & lightbox & 81.0 & 53.6 & 39.2 & 33.4 & 33.9 \\
     & sunlamp & 57.7 & 16.0 & 3.4 & 0.8 & 0.0 \\
    \bottomrule
  \end{tabular}
\end{table}

\begin{table}[ht]
  \caption{Coverage (\%) of the marginal 90\% region by range bin (m), translation, range-normalised, as the mean over the seeds of each arm. Bins are quintiles of the calibration split's range.}
  \label{tab:rangebins-translation-relative}
  \centering
  \scriptsize
  \setlength{\tabcolsep}{4pt}
  \begin{tabular}{l l ccccc}
    \toprule
    Arm & Domain & $<3.8$ & 3.8--5.3 & 5.3--6.7 & 6.7--8.3 & $\geq8.3$ \\
    \midrule
    frozen, no aug. & in-domain & 77.3 & 93.5 & 89.0 & 90.7 & 95.7 \\
     & lightbox & 38.7 & 56.2 & 50.1 & 48.5 & 54.0 \\
     & sunlamp & 31.8 & 39.6 & 21.9 & 9.7 & 5.3 \\
    \midrule
    frozen, full stack & in-domain & 74.5 & 88.5 & 88.6 & 94.3 & 92.9 \\
     & lightbox & 66.3 & 69.6 & 60.1 & 66.6 & 63.1 \\
     & sunlamp & 45.0 & 22.2 & 9.8 & 4.0 & 0.6 \\
    \midrule
    layer 4, no aug. & in-domain & 73.9 & 91.9 & 95.8 & 96.4 & 96.4 \\
     & lightbox & 35.1 & 33.7 & 31.1 & 32.3 & 38.3 \\
     & sunlamp & 35.0 & 18.0 & 4.6 & 1.0 & 0.2 \\
    \midrule
    layer 4, full stack & in-domain & 67.7 & 89.5 & 96.0 & 97.4 & 98.3 \\
     & lightbox & 57.0 & 48.3 & 42.8 & 41.9 & 47.0 \\
     & sunlamp & 33.7 & 12.3 & 3.6 & 1.0 & 0.1 \\
    \bottomrule
  \end{tabular}
\end{table}

\begin{table}[ht]
  \caption{Coverage (\%) of the marginal 90\% region by range bin (m), rotation, as the mean over the seeds of each arm. Bins are quintiles of the calibration split's range.}
  \label{tab:rangebins-rotation}
  \centering
  \scriptsize
  \setlength{\tabcolsep}{4pt}
  \begin{tabular}{l l ccccc}
    \toprule
    Arm & Domain & $<3.8$ & 3.8--5.3 & 5.3--6.7 & 6.7--8.3 & $\geq8.3$ \\
    \midrule
    frozen, no aug. & in-domain & 96.1 & 91.4 & 93.6 & 87.3 & 83.3 \\
     & lightbox & 75.9 & 74.2 & 71.1 & 69.3 & 64.2 \\
     & sunlamp & 64.2 & 56.5 & 61.4 & 56.2 & 52.5 \\
    \midrule
    frozen, full stack & in-domain & 92.0 & 92.8 & 92.9 & 90.1 & 84.9 \\
     & lightbox & 78.9 & 78.2 & 76.0 & 73.5 & 71.4 \\
     & sunlamp & 74.2 & 70.3 & 70.0 & 70.0 & 67.1 \\
    \midrule
    layer 4, no aug. & in-domain & 86.0 & 93.9 & 95.0 & 91.7 & 81.8 \\
     & lightbox & 22.8 & 29.0 & 27.5 & 24.5 & 19.7 \\
     & sunlamp & 11.1 & 11.6 & 11.3 & 10.3 & 7.1 \\
    \midrule
    layer 4, full stack & in-domain & 86.2 & 95.5 & 95.4 & 92.8 & 87.7 \\
     & lightbox & 35.6 & 40.9 & 38.3 & 34.5 & 29.5 \\
     & sunlamp & 17.3 & 17.7 & 16.8 & 11.0 & 9.3 \\
    \bottomrule
  \end{tabular}
\end{table}

\clearpage
\section{Reproducibility}
\label{app:repro}

\paragraph{Code, results and predictions.} Everything is in the public repository,
\url{https://github.com/lokindoo/spacecraft-pose-estimation}. The tag \texttt{arxiv-v1}
reproduces this version, and the tag \texttt{sim2science-2026-camera-ready} the workshop
version. Zenodo archives every tag under
\href{https://doi.org/10.5281/zenodo.24011140}{doi:10.5281/zenodo.24011140}, which lists each
version, and the workshop version alone under
\href{https://doi.org/10.5281/zenodo.24011141}{doi:10.5281/zenodo.24011141}. Each tag holds its
paper's source, with a script that checks every number in the paper against the repository's
\texttt{results/}. Per-image predictions for all 20 runs
behind the tables are released as CSV, so a new metric needs neither the data pipeline nor a
GPU.

\paragraph{Runs.} Configurations, seeds and the selection rule are as in
Sec.~\ref{sec:setup}. The eight published runs (seed 42 of both frozen arms, and all six
layer-4 runs) were trained on a free-tier Colab GPU. Frozen augmented seeds 43 and 44,
ablation arms A and B (three seeds each) and the control were trained with the released
package on an Apple M5 GPU. The frozen-backbone heads were trained on cached features on the
same machine's CPU, in minutes each.
On Colab a 20-epoch run took about 1.8\,h for the frozen un-augmented arm and 2.0--2.3\,h for
the layer-4 un-augmented arm. With augmentation, epochs took about 790\,s, some 4.4\,h per
run, whether or not the backbone was frozen. On the M5 most runs took 3.1--4.2\,h, and two
arm-A runs took about 14\,h of wall-clock time. A head on cached features trains in about
2.5 minutes.

\paragraph{Checkpoints.} \texttt{results/checkpoints.json} records, for each of the 17 trained
checkpoints, its configuration, seed, selected epoch, hardware and sha256. The weights are on
Zenodo with a model card, under CC~BY~4.0:
\href{https://doi.org/10.5281/zenodo.24140385}{doi:10.5281/zenodo.24140385}.

\paragraph{Software.} \texttt{requirements.txt} pins the versions the pipeline was written
against (PyTorch 2.13.0, torchvision 0.28.0). The M5 runs used Python 3.14.8, PyTorch 2.14.1
and torchvision 0.29.1, frozen in \texttt{requirements-m5.txt}. The Colab runs used the Colab
image of July 2026, whose versions were not recorded. The control rerun
(Sec.~\ref{sec:directional}) bounds what hardware and software together moved seed 42, and
none of the 81 unit tests needs data.

\section{Pre-registration record}
\label{app:prereg}

Every experiment added after review had its prediction and decision rule committed to version
control before its result was opened. The two files are published verbatim in the
repository's \texttt{preregistration/}, with every later edit and its time. The revision
predictions were committed on 28 Oct 2026 at 17:42 UTC. The extension's decision rules
followed on 30 Oct at 11:31 and its predictions at 11:54, and its results were opened at 13:36
the same day. These are commit times in the author's own repository, not a third-party
registration.

\begin{table}[ht]
  \caption{Pre-registered decision rules and their outcomes. Differences are paired by
  seed and given in rad, and ``3/3'' means all three seeds share the sign.}
  \label{tab:prereg}
  \centering
  \scriptsize
  \setlength{\tabcolsep}{3pt}
  \begin{tabular}{p{0.2\linewidth} p{0.38\linewidth} p{0.36\linewidth}}
    \toprule
    Experiment & Rule & Outcome \\
    \midrule
    \multicolumn{3}{@{}l}{\emph{Revision (locked 28 Oct)}} \\
    Arm A & Sunlamp, A $-$ full stack, 3/3 negative: keep ``blur and noise work against sunlamp outright'' & 2/3 ($-0.184$, $+0.006$, $-0.117$): softened to a hypothesis \\
    Arm B & Sunlamp, B $-$ A, 3/3 negative: the glare sentence gains evidence & 1/3 ($-0.018$, $+0.077$, $+0.072$): softened \\
    Control & Within 2 s.d.\ of published seed 42 on both real domains & Lightbox $+0.006$ ($\pm0.020$), sunlamp $-0.142$ ($\pm0.239$): inside; pairing holds \\
    Frozen seeds & Sunlamp s.d.\ $\geq 0.025$: ``only 0.03\,rad'' becomes ``no detectable change'' & s.d.\ 0.031: reworded \\
    Decomposition & One factor $\geq 2/3$ of the synthetic gain: name it; either factor moves sunlamp by $>2$ frozen s.d.: split the claim & Data 0.33, features 0.29 (53\%); largest sunlamp move 0.024 against 0.062: neither \\
    Head sweep & Beats published seed 42 by $>2$ frozen s.d.: ``under-trained''; recovers $\geq 50\%$ of the layer-4 gain on a real domain: restate & 0.913 against 1.053: tuned head reported; real domains unmoved: finding stands \\
    Synthetic-BN head & Share of the layer-4 gain recovered: $\geq 50\%$ / 0--50\% / $\leq 0$ & $-0.51$ (lightbox), $-0.27$ (sunlamp): ``not explained by normalisation statistics alone'' \\
    AdaBN & Improved if 6/6 checkpoints share the sign; coverage $<80\%$: ``coverage stays low'' & $\Erot$ 6/6 on both domains; $\Etrans$ 5/6 lightbox, 6/6 sunlamp; coverage at most 40.4\% \\
    Image statistics & Sunlamp clipped more than the full stack, and sharper than lightbox and the full stack; else two passages become hypotheses & Clipping holds (2.78\% vs 1.96\%); edge density fails (2.0\% vs 6.6\%, 10.7\%): passages reworded \\
    \midrule
    \multicolumn{3}{@{}l}{\emph{Extension (locked 30 Oct)}} \\
    Frozen + aug & Synthetic aug $-$ no aug, 3/3 positive: keep ``the reference degrades''; sunlamp 3/3 negative and larger than lightbox: qualify the directional claim & $+0.169$, $+0.161$, $+0.134$: kept; sunlamp $-0.128$, $-0.069$, $-0.132$, larger than lightbox on every seed: qualified \\
    Error shape & Near-$\pi$ share $\geq 0.28$ on all six: bimodal; body cone $\geq 0.18$ on all six and camera not: flips; reverse: in-plane & Near-$\pi$ 0.03--0.10: not bimodal; body cone $\geq 0.18$ on 2/6 per domain: neither \\
    Range bins & Both in-domain translation patterns hold and the extreme bin excludes 90\%: keep; monotone but intervals include 90\%: soften; not monotone: cut & Ambiguous: ``monotone'' undefined. Near-to-far, both hold and the extreme bin excludes 90\% on 7/8 and 8/8; strictly over five bins, 3/8 and 2/8. Reported as near and far numbers \\
    Ablation coverage & Coverage opposite to error on $\geq 5/6$ per domain: ``coverage mirrors error'' & 4/6 on each domain: not claimed \\
    \bottomrule
  \end{tabular}
\end{table}

\paragraph{Deviations and corrections.} Frozen augmented seeds 43 and 44 were pre-registered
for this version but finished after the camera-ready's results freeze. Once their outcome was
known they were added to the camera-ready anyway, the only result to cross the freeze. The
range-bin rule also left ``monotone'' undefined, as Table~\ref{tab:prereg} notes.

\paragraph{Not pre-registered.} Three results in the body were found or computed outside the
pre-registration: the transfer plot (Fig.~\ref{fig:transfer}), AdaBN's effect on translation (Sec.~\ref{sec:bn}) and the collapse
of sunlamp translation coverage with range (Sec.~\ref{sec:conformal}). Every point in the
transfer plot had already been published.

\end{document}